%% file: main.tex
\documentclass[conference]{IEEEtran}
\usepackage{bm}
\usepackage{times}
\usepackage{xcolor}
\usepackage[numbers]{natbib}
\usepackage{multicol}
\usepackage{amsmath}
\usepackage{amssymb}
\usepackage{algorithm}
\usepackage{algpseudocode}
\usepackage[pdftex]{graphicx}
\usepackage{caption}
\usepackage[bookmarks=true,linkcolor=blue,colorlinks = true, citecolor=orange]{hyperref}
\usepackage{float}
\usepackage{booktabs}
\usepackage{graphics}
\usepackage{placeins}
\usepackage{afterpage}
\usepackage{multirow}
\usepackage{pifont}
\usepackage[table]{xcolor}
\usepackage{verbatim}

\definecolor{lightgray}{rgb}{0.5, 0.5, 0.5}
\newcommand{\cmark}{\ding{51}}
\newcommand{\xmark}{\ding{55}}

\begin{document}

\title{Automated Synthesis of Facial Mechanisms for Conversational Animatronic Robots}

\newcommand\blfootnote[1]{%
  \begingroup
  \renewcommand\thefootnote{}\footnote{#1}%
  \addtocounter{footnote}{-1}%
  \endgroup
}
\author{Zongzheng Zhang$^{*1, 2}$, Zi Lin$^{*1}$, Jiawen Yang$^{1}$, Ziqiao Peng$^{1}$, \\Junyan Lao$^{1}$,  Lin Cheng$^{4}$, Huazhe Xu$^{3}$, Hang Zhao$^{3}$, Hao Zhao\textdagger$^{1, 2}$ \vspace{0.2cm}\\
 $^1$ Institute for AI Industry Research (AIR), Tsinghua University  $^2$ Beijing Academy of Artificial Intelligence (BAAI) \\
 $^3$ Institute for Interdisciplinary Information Sciences (IIIS), Tsinghua University 
 $^4$ Beihang University \\
 $^*$ Equal contribution  \textdagger Corresponding author\\
 \href{https://zzongzheng0918.github.io/automated-facial-mechanisms-synthesis/}{\textcolor{orange}{https://zzongzheng0918.github.io/automated-facial-mechanisms-synthesis/}}
}




%

\makeatletter
\let\@oldmaketitle\@maketitle
\renewcommand{\@maketitle}{\@oldmaketitle
 \begin{center}
  \captionsetup{type=figure}
    \includegraphics[width=1.0\textwidth]{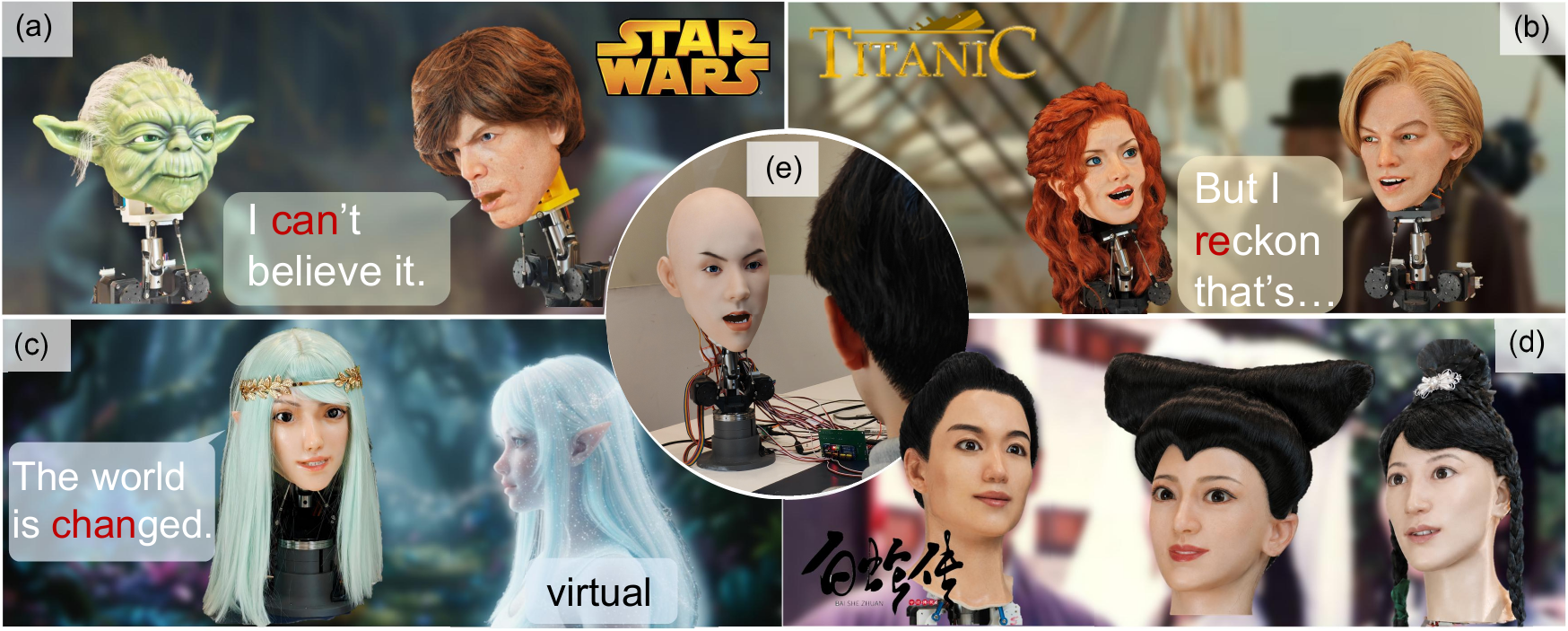}
    \caption{We demonstrate our end-to-end physical conversational face system across diverse multi-round interactions: (a) reenactments of \textit{Star Wars} (Yoda–Luke Skywalker) and (b) \textit{Titanic} (Rose–Jack); (c) a dialogue between a real elf and a virtual elf; (d) a three-character encounter from the Chinese folktale \textit{Legend of the White Snake}; and (e) human–robot dialogue.
}
\vspace{-0.5cm}
\label{fig:teaser}
\end{center}
}
\makeatother

\maketitle
\setcounter{figure}{1}

\input{sec/0_abs}
\IEEEpeerreviewmaketitle
\input{sec/1_intro}

\input{sec/2_related_work}

\input{sec/3_method1}

\input{sec/3_method2}
\input{sec/4_experiment}
\input{sec/6_conclusion}

\bibliographystyle{plainnat}
\bibliography{main}

\input{sec/X_supp}

\end{document}

%% file: sec/0_abs.tex
\begin{abstract}
Animatronic faces are a central component of socially interactive robots, enabling rich nonverbal communication through facial articulation. However, state-of-the-art animatronic faces are typically tailored systems: each new facial geometry requires extensive manual mechanical redesign, making large-scale personalization prohibitively slow and costly. In this work, we pursue automated and scalable mechanical face synthesis, aiming to rapidly generate a physically realizable facial mechanism for a wide range of facial geometries. We introduce a parametric, linkage-driven mechanical face template whose topology and actuator layout are explicitly parameterized to support systematic scaling and retargeting across diverse facial morphologies. Building on this template, we propose a hierarchical automatic design algorithm that takes a single 2D portrait as input, reconstructs a target 3D face, and synthesizes a collision-free, manufacturable internal mechanism. The algorithm combines anatomy-guided feasible motion volumes, Action Unit (AU)-derived trajectory-based expressiveness objectives, and a collision-driven outer-loop refinement strategy. Beyond hardware synthesis, we argue that future mechanical faces deployed at scale must engage in bidirectional, multi-turn conversation, rather than functioning solely as speaking or listening heads. To this end, we develop a dual-identity conversational facial motion synthesis framework that jointly models speaking and listening behaviors from audio, producing temporally coherent 3D facial motion suitable for physical execution. We validate our system through extensive experiments, including (i) quantitative evaluation of automatic mechanism synthesis across diverse facial geometries, (ii) comparisons against manual mechanical design, (iii) benchmarks on conversational facial motion synthesis and real-time deployment, and (iv) perceptual user studies.
\end{abstract}

%% file: sec/1_intro.tex
\section{Introduction}
\label{sec:intro}

Imagine a future where thousands of robots coexist with humans, each endowed with a distinct face, identity, and personality. Rather than sharing a single canonical appearance, these robots would exhibit thousands of unique mechanical faces (from realistic human likenesses to stylized fictional characters) capable of engaging in rich, multi-turn conversations. As illustrated in Fig.~\ref{fig:teaser}, such animatronic faces would not merely speak, but participate in bidirectional dialogue, react as listeners, and sustain expressive interactions across diverse narratives and social contexts. Realizing this vision of massively personalized, conversational mechanical faces represents a foundational step toward socially integrated robotics.

However, despite decades of progress in animatronic face design~\cite{kobayashi1993study, kobayashi2000study, chen2021smileeva}, today’s state-of-the-art systems remain fundamentally tailored~\cite{hu2024human, zhu2025awakening, zhang2025morpheus}. High-fidelity platforms achieve impressive realism through carefully engineered actuation layouts, transmission systems, and handcrafted linkages~\cite{liu2022real, fan2025soft, li2024driving}. Yet each of these faces is designed for a single fixed geometry. Adapting such systems to a new facial morphology typically requires extensive manual mechanical redesign, iterative prototyping, and expert intervention. As a result, current approaches scale poorly: they cannot feasibly support the rapid creation of hundreds or thousands of mechanically distinct faces, let alone support the diverse characters demonstrated in Fig.~\ref{fig:teaser}. This reliance on custom mechanical design forms a critical bottleneck that prevents animatronic faces from scaling beyond isolated demonstrations.

To overcome this limitation, we argue that animatronic faces must be treated not as handcrafted artifacts, but as automatically synthesizable systems. Our core design principle is simple: instead of redesigning mechanisms from scratch for each face, we start from a parametric mechanical face template whose topology, actuator layout, and kinematic structure are explicitly parameterized. Given this template, we seek an automated pipeline that can rapidly adapt the internal mechanism to any given facial geometry, producing a collision-free, manufacturable design with minimal human intervention. In this paradigm, mechanical face design becomes a computational problem~\cite{du2016computational, coros2013computational, thomaszewski2014computational, kodnongbua2022computationalgripper}.

Building on this principle, we introduce an end-to-end automated mechanical face synthesis pipeline. Starting from a single 2D portrait, our system reconstructs a 3D facial geometry and initializes a modular linkage-driven face template. We then perform hierarchical optimization to adapt the template to the target face. \textbf{At the module level}, we synthesize kinematic structures that maximize expressiveness under anatomy-guided feasible motion volumes and AU-derived trajectory objectives. \textbf{At the assembly level}, we resolve spatial conflicts using a collision-driven outer-loop refinement strategy that iteratively adjusts base poses or schedules expression amplitudes. This hierarchical design tightly couples kinematic synthesis with engineering validation, enabling the automatic generation of high-DoF facial mechanisms that are both expressive and physically realizable. As demonstrated in Fig. ~\ref{fig:teaser}, this approach generalizes across human, stylized, and non-human faces, including characters with drastically different morphologies.

Crucially, scalable hardware alone is not sufficient. If animatronic faces are to be deployed at scale in human environments, they must function as conversational agents, not merely as talking heads. Yet most existing facial motion generation and mapping methods are designed for a single face and single-speaker setting~\cite{zhu2025awakening, zhang2025morpheus, hu2026learning,xu2026singingbot,li2024driving}. They generate speech-driven expressions without modeling listening behaviors or multi-turn interaction, and are often ill-suited for physical execution. In contrast, we posit that future mechanical faces should participate in dialogue, alternating naturally between speaking and listening roles. To this end, we introduce a dual-identity conversational facial motion synthesis framework that jointly models speaking and listening behaviors from audio, producing temporally coherent 3D facial motion suitable for real-time robotic actuation. We further develop a semantic, region-wise mapping strategy that robustly translates synthesized facial motion into actuator commands, bridging the gap between digital animation and physical execution.

We validate our approach through extensive experiments spanning both hardware and software. \textbf{On the mechanical side}, we quantitatively evaluate the success rate, convergence speed, and expressiveness of our automated design pipeline across diverse facial geometries, and compare against manual design and alternative optimization strategies. \textbf{On the interaction side}, we benchmark our conversational facial motion synthesis against state-of-the-art talking-head methods in both speaker and listener roles, and demonstrate real-time deployment on physical robots. \textbf{Finally}, perceptual user studies confirm that our end-to-end system delivers more natural, engaging, and socially convincing interactions. Together, these results demonstrate that our framework provides a scalable and practical path toward the vision of thousands of personalized, conversational animatronic faces.

%% file: sec/2_related_work.tex
\section{Related Work}
\label{sec:related work}

\subsection{Mechanical Face Platform}
The evolution of mechanical face design has transitioned from early exploration of actuation modalities to today's hyper-realistic platforms. Early research established the feasibility of FACS-based motion using pneumatic muscles and micro-actuators~\cite{kobayashi1993study,kobayashi2000study,kobayashi2003realization,hashimoto2006development,hashimoto2008dynamic,berns2006control,itoh2006mechanical,oh2006design}, later incorporating bio-inspired materials and tendon-driven mechanisms to improve compliance~\cite{hanson2002identity,weiguo2004development,habib2014learning,fortunati2021rise,li2024driving}. Subsequent efforts focused on system integration, embedding dense DoFs into full-body humanoids~\cite{kaneko2009cybernetic,nakaoka2009creating,lee2008development,ahn2012designing,lin2009realization} and android heads~\cite{habib2014learning,fortunati2021rise,glas2016erica}, while investigating social engagement through developmental platforms like Cog and Affetto~\cite{brooks1998cog,lutkebohle2010bielefeld,ishihara2011realistic,allison2008design}. Recently, the field has shifted toward high-fidelity architectures with refined transmission systems, exemplified by platforms such as Abel, Ameca, Nikola, and EMO~\cite{cominelli2021abel,nieto2025robot,faraj2021facially,chen2021smileeva,hu2024human,yang2022optimizing,yang2025hapi}. Latest designs further explore diverse points in the design space, from robust hybrid actuation schemes (Morpheus) to low-cost soft solutions~\cite{sheng2025review,liu2022real,fan2025soft,zhu2025awakening,zhang2025morpheus}. Despite these advancements, a critical limitation persists: state-of-the-art systems are bespoke designs engineered for a single fixed geometry. Adapting these sophisticated architectures to new 3D faces typically necessitates substantial manual redesign, a scalability bottleneck our work aims to resolve.

\subsection{Automatic Robot Design}
Computer graphics has extensively explored synthesizing mechanisms from motion specifications, ranging from planar linkages and characters~\cite{coros2013computational, bacher2015linkedit, thomaszewski2014computational} to automata derived from motion capture~\cite{ceylan2013designing, zhu2012motion}. Complementary tools have facilitated rapid fabrication~\cite{megaro2014chacra, cali20123d} and template retargeting~\cite{zhang2017functionality}, though these methods typically prioritize kinematic feasibility over the internal interference required for physical operation. In robotics, optimization extends to dynamics, with task-based methods tuning cable-driven mechanisms~\cite{li2017task, kawaharazuka2023design, islam2024task, penta2016analysis}. Furthermore, morphology-control co-optimization utilizing differentiable, RL, and graph-based frameworks~\cite{xu2021end, ha2019reinforcement, schaff2019jointly, he2024morph, wang2019neural, zhao2020robogrammar, kawaharazuka2024robot, li2023modular, sun2026knowledge} has been applied across domains including locomotion~\cite{geilinger2018skaterbots, ha2016taskleg}, soft robotics~\cite{lipson2000automatic, hiller2011automatic}, manipulation~\cite{mannam2024design, gilday2025embodied, gilday2024exploiting, yi2025codesign, kodnongbua2022computationalgripper}, and multicopters~\cite{du2016computational, xu2019learning}. However, these strategies often yield coarse topologies, lacking the precision to embed complex linkages into constrained facial volumes. Specific to facial robotics, seminal works~\cite{bickel2012physical, huber2021designing} address the coupling of rigid actuators and deformable skin. Yet, they treat the kinematic structure as fixed, optimizing only placement. In contrast, our framework unifies kinematic synthesis with engineering validation to automate the design of high-DOF facial mechanisms.

\subsection{Talking Head Generation and Expression Mapping}

While 2D audio-driven generation offers visual fidelity~\cite{prajwal2020lip, zhang2023sadtalker, kong2024hunyuanvideo, peng2025omnisync, lin2025omnihuman, zhao2020learning}, its high latency and limited perspective robustness~\cite{zhou2025interactive,zhu2025infp, xu2026singingbot, hu2026learning, zhang2025shouldershot} make 3D synthesis preferable for robotic actuation~\cite{cudeiro2019capture, richard2021meshtalk,chu2025artalk, zhang2025morpheus}. 3D synthesis has evolved from deterministic lip-sync~\cite{fan2022faceformer, xing2023codetalker, li2023understanding} to emotional expressiveness~\cite{peng2023emotalk, danvevcek2023emotional, liu2024emoface} and cross-identity generalization~\cite{lu2025lsf, fan2024unitalker, chen2025cafe}. Crucially, recent research shifts from monologue~\cite{peng2023selftalk, peng2024synctalk, peng2025synctalk++} to dyadic interaction, employing dual-turn frameworks to model reciprocal dynamics and listening behaviors~\cite{ peng2025dualtalk, ng2024audio, tran2024dim, chu2025unils, chen2025towards}. To bridge the gap to physical robots, mapping approaches have progressed from geometric correspondence~\cite{chen2021smileeva, fan2025soft, wu2024retargeting, heisler2025iphone, zhu2025awakening} to semantic representations and data-driven optimization~\cite{liu2024unlocking, li2025x2c, zhang2025exface, zhang2025fabg, yang2025hapi, li2024driving}. Unlike prior works restricted to single-speaker generation or frame-wise mapping, we strictly enforce a joint modeling of dual-turn synthesis and semantic region-wise mapping to enable robust multi-turn conversational interaction.

%% file: sec/3_method1.tex
\section{Automated Scalable Mechanism Design}
\label{sec: method1}
In this section, we present an end-to-end pipeline for \textbf{automated hardware design} of a scalable robotic face. We first introduce a \textbf{scalable hardware template} derived from a canonical mean face model, where the actuator layout and linkage topology are parameterized to support rapid adaptation across diverse facial morphologies (Sec.~\ref{sec:Face Template Design}). Building on this template, we then propose a \textbf{hierarchical automatic design algorithm} (Algorithm~\ref{alg:overall optimization}) that takes a single 2D portrait as input, reconstructs the target 3D facial geometry, and progressively optimizes the template parameters under manufacturability and non-interference constraints (Sec.~\ref{sec:automatic design algorithm}).

\subsection{Parametric Face Template Design}
\label{sec:Face Template Design}
\begin{figure}
    \centering
    \includegraphics[width=1.0\linewidth]{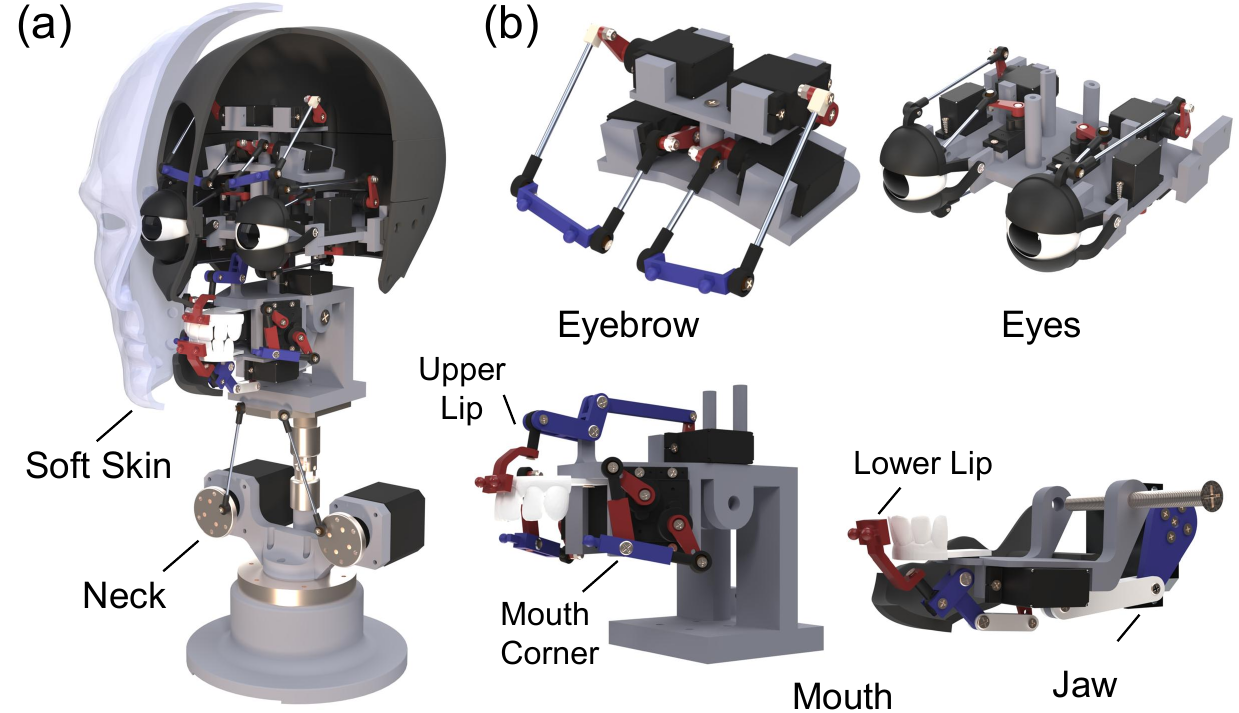}
    \caption{\textbf{Mechanical face template.} (a) Full assembly of the linkage-driven robotic face with a soft skin and a 3-DoF neck module. (b) Exploded view of the four modular facial mechanisms—eyebrow, eyes, mouth, and jaw.}
   \vspace{-0.8cm}
    \label{fig:hardware template}
\end{figure}

We develop a modular, parametric mechanical-face template that serves as the starting point of our automated design pipeline.  The face template is composed of four modules: \textbf{eyebrow}, \textbf{eyes}, \textbf{mouth}, and \textbf{jaw} (Fig.~\ref{fig:hardware template}). 
The eyebrow module contains two kinds of motion: the vertical movement of the eyebrow ridge and the brow center. Both are realized by a spatial six-bar mechanism with $2$ DoF.
The eye module contains $8$ DoF. It comprises four eyelid linkages (upper/lower lids for both eyes) and two rotational DoFs per eyeball.
The mouth module integrates mechanisms for the upper/lower lips and mouth corners.  Both the upper and lower lips are actuated by spatial four-bar linkages whose end-effector motions follow vertical trajectories to match the predominant anatomical direction of lip deformation. The mouth corners are driven by a five-bar mechanism that supports three principal motion modes: upward raising, lateral translation, and downward pulling. The jaw module is actuated by a four-bar mechanism.
Finally, head pose is produced by a neck module consisting of a $2$\,DoF parallel mechanism for nodding/tilting and an additional rotational DoF for yaw.

\subsection{Hierarchical Automatic Design}
\label{sec:automatic design algorithm}

\begin{figure*}
    \centering
    \includegraphics[width=1.0\linewidth]{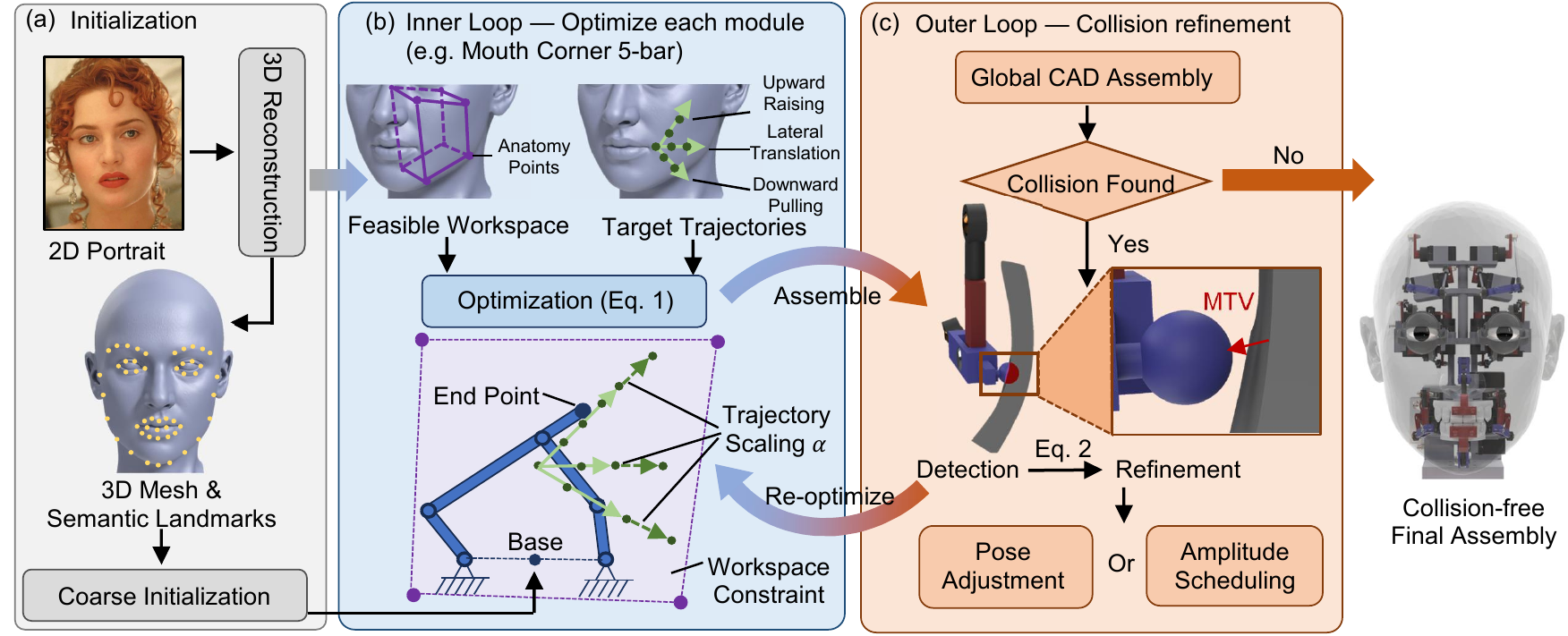}
    \caption{\textbf{Overview of the hierarchical automatic design pipeline.} (a) \textbf{Initialization}: From a 2D portrait, we reconstruct a 3D head mesh, semantic landmarks, and initialize module base poses. (b) \textbf{Inner loop}: We perform module-wise kinematic synthesis under anatomy-guided feasible volumes and AU-derived trajectories to maximize expressiveness (illustrated with the mouth-corner five-bar mechanism). (c) \textbf{Outer loop}: We assemble the global CAD model, detect interferences, and apply MTV-guided base-pose updates (or, if necessary, reduce the amplitude limit) until a collision-free assembly is obtained.}
   \vspace{-0.6cm}
    \label{fig:automatic mechanical design}
\end{figure*}

\subsubsection{Coarse initialization}
Given an input RGB portrait image $I$, we use the framework~\cite{zielonka22mica} to construct a normalized 3D facial mesh $\mathcal{M}$ together with a set of semantically labeled 3D facial landmarks $\mathbf{P}$. We then import $\mathcal{M}$ into a CAD environment and apply a scaling factor to bring the head geometry into a physically plausible metric range. These 3D semantic points are then used for the subsequent module-level coarse initialization (Fig.~\ref{fig:automatic mechanical design}(a)).

Using the scaled semantic landmarks, we perform a module-level coarse initialization to obtain the initial base poses $\mathbf{p}^{\mathrm{base}}$. The \textbf{eyes} and \textbf{jaw} modules are fully determined at this stage. Specifically, the eye geometry is directly computed by peri-ocular landmark subsets, and the jaw module is anchored by the jawline landmarks, yielding a fixed initialization. In contrast, the \textbf{eyebrow} and \textbf{mouth} (mouth corners and lips) modules require further kinematic synthesis. We initialize their base poses $\mathbf{p}^{\mathrm{base}}_{\mathrm{brow}}$ and $\mathbf{p}^{\mathrm{base}}_{\mathrm{mouth}}$ by scaling the template anthropometric offsets according to the reconstructed head scale:
$\mathbf{p}^{\mathrm{base}}_{k} \leftarrow \mathbf{p}^{\mathrm{base}}_{k,0} + \beta\,\Delta \mathbf{p}^{\mathrm{tmpl}}_{k},$
where $\Delta \mathbf{p}^{\mathrm{tmpl}}_{k}$ denotes the corresponding relative offsets in the mean-face template, and $\beta$ is the global scaling factor applied to $\mathcal{M}$. 
Their structural parameters are initialized with the template values, which serve as the starting point for the module-wise kinematic design optimization in the next step (see App.~\ref{sec: app-coarse alignment} for details).

\subsubsection{Module-wise kinematic design optimization}
\label{sec: module-wise optimization}
With the rigid modules (eyes and jaw) geometrically determined in the coarse initialization phase, this step focuses on the kinematic synthesis of the eyebrow and mouth modules. The core challenge lies in synthesizing mechanism parameters that maximize expressive amplitude while strictly adhering to the anatomical spatial constraints imposed by the facial skin.

\textbf{Anatomy-guided feasible motion volumes.} To ensure the synthesized mechanisms operate within bio-plausible limits, we define the feasible workspace  $\Omega_\mathrm{feas}$ for the eyebrow, mouth corner, and lip regions (Fig.~\ref{fig:automatic mechanical design}(b); details in App.~\ref{sec: app-Determine local 3D design space.}), following the evidence from~\cite{drake2015grays, chong2021three}. We construct $\Omega_\mathrm{feas}$ as the maximal anatomically valid envelope, so that the optimizer retains sufficient freedom to maximize expressiveness. Simultaneously, strictly confining the solution to this valid manifold acts as a regularization term. This accelerates convergence by pruning the search space of physically impossible poses and mitigates the risk of the solver getting trapped in invalid local minima, thereby guiding the optimization toward realizable solutions.

\textbf{Trajectory-based expressiveness objectives.} We derive module-specific motion trajectories from facial Action Units (AUs)~\cite{ekman1978facial} to ensure that the designed mechanisms follow anatomically consistent patterns of human facial motion. Concretely, we define a set of canonical target trajectories $\mathcal{B} = \{ \mathbf{T}_1, \dots, \mathbf{T}_K \}$, where each $\mathbf{T}_k$ is represented by a sequence of $M$ sampled points in the local workspace (details in App.~\ref{sec: app-Determine trajectory}). We reformulate the goal of maximizing expressiveness as a trajectory scaling problem. For each trajectory $\mathbf{T}_k$, we introduce a scalar amplitude coefficient $\alpha_k$. Crucially, this scaling is directional: rather than isotropically enlarging the mechanism, it linearly extrapolates the waypoints along the trajectory path $\mathbf{T}_k$ (Fig.~\ref{fig:automatic mechanical design}(b)). By optimizing for larger $\{\alpha_k\}_{k=1}^K$ values, we aim to synthesize mechanisms capable of executing high-amplitude expressions while strictly preserving the semantic direction of the original human movements.

\textbf{Optimization formulation.} For each mechanical structure $k$, we perform kinematic design optimization in its local coordinate frame. Specifically, using the module base pose $\mathbf{p}_{base}$ from coarse initialization, we transform $\mathbf{T}_k$ to each local frame. We formulate the module-wise mechanism synthesis as a constrained non-linear optimization problem. Given the target trajectories $\mathbf{T}_k$, we seek to determine the optimal structural parameters $\mathbf{\Phi}_k$ (link lengths), the set of discrete joint configurations $\mathbf{\Theta}_k$, and the expression amplitude coefficients $\mathbf{\alpha}_k$. We define a unified cost function $\mathcal{L}$ to balance expressive range, tracking accuracy, and kinematic manipulability:
\begin{equation}
\begin{aligned}
\min_{\mathbf{\Phi},\,\mathbf{\Theta},\,\boldsymbol{\alpha}}\quad 
& \mathcal{L}
= \omega_\mathrm{amp}\mathcal{L}_\mathrm{amp}+\omega_\mathrm{fit}\mathcal{L}_\mathrm{fit}-\omega_\mathrm{man}\mathcal{L}_\mathrm{man} \\
\text{s.t.}\quad 
& \mathbf{P}(\boldsymbol{\theta}_{k,m}) \subset \Omega_{\mathrm{feas}}, 
\quad \forall m\in\{1,\dots,M\},\\
& 0<\alpha_k \le \alpha^\mathrm{limit}_{k}, 
\quad \forall k\in\{1,\dots,K\},
\end{aligned}
\label{eq:innerloop_overall_obj_constrained}
\end{equation}
where $\omega_\mathrm{amp}$, $\omega_\mathrm{fit}$, and $\omega_\mathrm{man}$ are user-defined weighting factors. The individual objective terms are defined as follows: $\mathcal{L}_{\mathrm{amp}} = 1/\alpha_k$; $\mathcal{L}_{\mathrm{fit}} = \sum_{m=1}^{M} \| f(\boldsymbol{\theta}_{k,m},\mathbf{\Phi})-\mathbf{p}_{k,m}(\alpha_k) \|_2^2$, where $\mathbf{p}_{k,m} \in \mathbf{T}_k$; $\mathcal{L}_{\mathrm{man}} = \sum_{m=1}^{M} \sqrt{\det(\mathbf{J}(\boldsymbol{\theta}_{k,m})\mathbf{J}(\boldsymbol{\theta}_{k,m})^{\top})}$. 
Here, $m$ denotes the waypoint index along the trajectory. $\boldsymbol{\theta}_{k,m}$ denotes the instantaneous joint configuration at step $m$, $f(\cdot)$ represents the forward kinematics, and $\mathbf{J}$ is the geometric Jacobian matrix utilized to maximize workspace manipulability (see App.~\ref{sec: app-kinematics analysis} for explicit derivations).

\begin{algorithm}[t]
\caption{Hierarchical Automatic Design}
\label{alg:overall optimization}
\small
\setlength{\emergencystretch}{2em}
\begin{algorithmic}[1]
\raggedright
\Require 2D portrait image $I$; template face design
\Ensure Final assembly $\mathcal{A}$

\State $(\mathcal{M},\mathbf{P}) \gets \Call{Reconstruct3D}{I}$
\State $\{\mathbf{p}^\mathrm{base}\} \gets \Call{CoarseInitialization}{\mathcal{M},\mathbf{P}}$ \textcolor{lightgray}{\Comment{Initialization}}

\For{iteration $s=1$ to $S$}
  \For{mechanism $k\in\{\mathrm{brow},\mathrm{mouth}\}$}  \textcolor{cyan!80}{\Comment{Inner loop}}
    \State $\{\mathbf{T}_k\}\gets \Call{GlobalToLocal}{\{\mathbf{T}_k\},\mathbf{p}_k^\mathrm{base}}$
    \State $(\mathbf{\Phi}_k,\mathbf{\Theta}_k,\boldsymbol{\alpha}_k)\gets
    \Call{SolveEq.~\ref{eq:innerloop_overall_obj_constrained}}{\Omega_{\mathrm{feas}},\{\mathbf{T}_k\},\{\alpha_k^\mathrm{limit}\}}$
  \EndFor
  \State $\alpha_k^\mathrm{limit}\gets \min(\alpha_k^\mathrm{limit},\alpha_k^{*}),\ \forall k$ \textcolor{orange!90!black}{\Comment{Outer loop}}

  \State $\mathcal{A}\gets \Call{AssembleCAD}{\{\mathbf{p}_k^\mathrm{base}\},\{\mathbf{\Phi}_k\}}$
  \State $\mathcal{C}\gets \Call{CollisionDetection}{\mathcal{A},\{\mathbf{\Theta}_k\}}$
  \If{$\mathcal{C}=\emptyset$} \State \textbf{break} \EndIf

  \For{each mechanism $k$ with constraints $\mathcal{C}_k$}
    \State $(\{\mathbf{n}_j,\delta_j\}_{j\in\mathcal{C}_k})\gets \Call{MTV}{\mathcal{C}_k}$
    \State $\Delta\mathbf{p}_k \gets
    \Call{SolveEq.~\ref{eq:outloop_optimization}}{\{\mathbf{n}_j,\delta_j\}_{j\in\mathcal{C}_k},\epsilon}$
    \If{$\|\Delta\mathbf{p}_k\|_2 > D_{\max}$} \textcolor{orange!90!black}{\Comment{Amplitude scheduling}}
  \State $\alpha_k^{\mathrm{limit}} \gets \gamma\,\alpha_k^{\mathrm{limit}}$ 
  \State \textbf{continue} \textcolor{orange!90!black}{\Comment{Skip update, re-optimization next iter}}
\EndIf
    \State $\mathbf{p}^\mathrm{base}_k \gets \mathbf{p}^\mathrm{base}_k + \Delta\mathbf{p}_k$ \textcolor{orange!90!black}{\Comment{Pose adjustment}}
\EndFor
\EndFor

\State \Return \ensuremath{\mathcal{A}}

\end{algorithmic}

\end{algorithm}

\subsubsection{Global CAD assembly and collision-driven refinement}
This step integrates all modules into a global CAD assembly via API and resolves spatial conflicts through a collision-driven outer-loop refinement strategy. For a mechanism $k$, we sample discrete configurations $\{ \boldsymbol{\theta}_{k,m} \}_{m=1}^M$ along its optimized trajectory and utilize a collision detection library FCL~\cite{pan2012fcl} to identify interferences between the $k$-th module and the environment (skull and adjacent mechanisms). For every active collision constraint $j \in \mathcal{C}_k$, we compute a \textbf{Minimum Translation Vector} $\mathbf{v}_\mathrm{int} = \delta_j \cdot \mathbf{n}_j$, where $\delta_j$ is the penetration depth and $\mathbf{n}_j$ is the separation normal (Fig.~\ref{fig:automatic mechanical design}(c)). To resolve multiple simultaneous collisions while minimally perturbing the design, we solve a QP problem:
\begin{equation}
\begin{aligned}
  \min_{\Delta \mathbf{p}_k} \quad & \|\Delta \mathbf{p}_k\|^2_2  \quad
\text{s.t.} \quad & \mathbf{n}_j^\top \Delta \mathbf{p}_k  \geq \delta_j + \epsilon, \quad \forall j \in \mathcal{C}_k, \\
\end{aligned}
\label{eq:outloop_optimization}
\end{equation}
where $\epsilon$ is a safety clearance margin. We accept the pose update only if $\Delta \mathbf{p}_k$ falls within a threshold $D_\mathrm{max}$; otherwise, we contract $\alpha^\mathrm{limit}_k$ via a decay factor $\gamma \in (0,1)$.
The updated $\mathbf{p}^\mathrm{base}_k$ is fed back to the inner-loop optimization phase (Sec.~\ref{sec: module-wise optimization}), where $\mathbf{\Phi}_k$ and $\mathbf{\Theta}_k$ are re-optimized in the new local frame. The outer loop iterates until all collisions are satisfied ($\mathcal{C} = \emptyset$), outputting the final assembly $\mathcal{A}$.

%% file: sec/3_method2.tex
\section{Interaction Synthesis and Control}
\label{sec: method2}

\begin{figure*}
    \centering
    \includegraphics[width=1.0\linewidth]{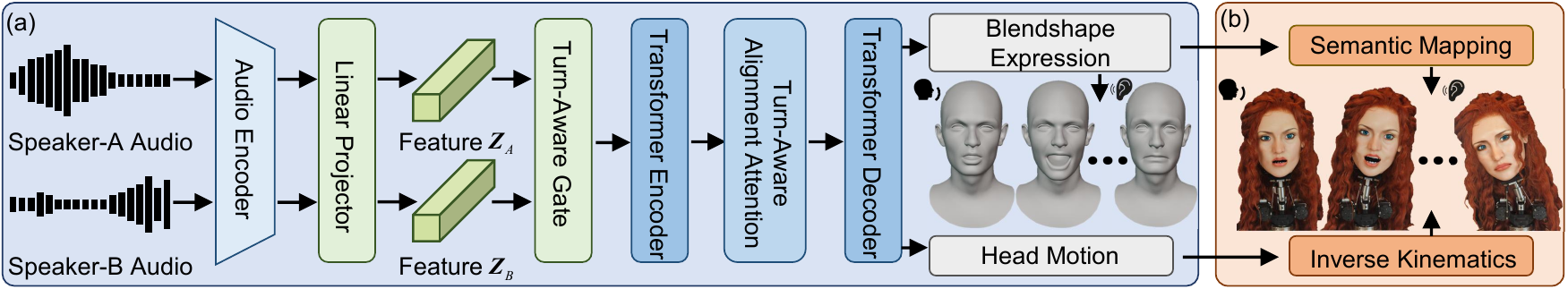}
    \caption{\textbf{Overview of interaction synthesis and control.} (a) Given dual-speaker audio, the talking-head model predicts multi-round facial motion for both speaking and listening. (b) The predicted coefficients are mapped to robot motor commands via region-wise regressors and neck IK, enabling physically feasible multi-round expressions on the mechanical face.}
   \vspace{-0.5cm}
    \label{fig:initeraction synthesis and mapping}
\end{figure*}

Following the automated mechanical-face adaptation described above, we present the algorithms that \textbf{drive interactive expressions on the physical robot}. We first generate \textbf{3D facial motion} for multi-round dialogue, modeling both speaking and listening behaviors under conversational context (Sec.~\ref{sec: 3d talking head synthesis}). We then \textbf{map} the synthesized digital facial motion to robot-actuable commands (Sec.~\ref{sec: mapping}). Together, these modules bridge conversational interaction, 3D facial animation, and embodied execution on the mechanical face.

\subsection{Interactive Talking Head Synthesis}
\label{sec: 3d talking head synthesis}

\textbf{Unified dual-speaker interaction framework.} The framework (Fig.~\ref{fig:initeraction synthesis and mapping}(a)) takes the audio from Speaker-A and Speaker-B as input and directly outputs a temporally coherent sequence of blendshape parameters that drives the synthesized facial motion. First, we use a pre-trained audio encoder Wav2Vec 2.0~\cite{baevski2020wav2vec} to map $\mathbf{A}_A \in \mathbb{R}^{T \times F}$ and $\mathbf{A}_B \in \mathbb{R}^{T \times F}$ into a shared latent space, where $T$ is the sequence length and $F$ is the sampling rate. We then apply a learnable linear projection $\mathbf{W}_a$ to align both streams to a shared feature dimension $d$:
\begin{equation}
    \mathbf{Z}_A = \mathbf{W}_a \big(E_{\mathrm{aud}}(\mathbf{A}_A)\big),\qquad
    \mathbf{Z}_B = \mathbf{W}_a \big(E_{\mathrm{aud}}(\mathbf{A}_B)\big) ,
\label{eq: audio encoder}
\end{equation}
where $\mathbf{Z}_A, \mathbf{Z}_B \in \mathbb{R}^{L \times d}$ serve as the unified audio feature. We add temporal positional encoding and style embeddings to get $\tilde{\mathbf{Z}}_A, \tilde{\mathbf{Z}}_B$. Since the two speakers are largely non-overlapping in speech, we regulate the contribution of the two audio streams via a turn-aware gate derived from speech activity. Let $g_A(t) \in [0, 1]$ indicate whether Speaker-A is speaking at frame $t$. We form a turn-conditioned audio memory for Speaker-A as: $\mathbf{M}_A(t)
= g_A(t)\,\tilde{\mathbf{Z}}_A(t)
+ \big(1-g_A(t)\big)\,\tilde{\mathbf{Z}}_B(t).$ Then we adopt a Transformer backbone. Specifically, an encoder models long-range dependencies of the turn-conditioned memory. To further stabilize speaking–listening transitions, we apply a turn-aware alignment attention. We decode the full motion using 
$T$ learnable motion queries with positional encodings. Finally, a regression head predicts Speaker-A blendshape parameters $\hat{\mathbf{Y}} \in \mathbb{R}^{T\times 55}$ (details in App.~\ref{sec: app-talking head synthesis}).

\textbf{Dataset construction.}
While recent conversational interaction datasets~\cite{peng2025dualtalk, chu2025unils} utilize FLAME~\cite{li2017learning} parameters, its expression coefficients are not aligned with semantically interpretable Blendshape~\cite{Lewis2014PracticeAT} controls, which complicates the mapping process. We therefore build a dataset of paired two-speaker audio and per-frame blendshape coefficients. 

We collect raw conversation videos from YouTube, and RealTalk ~\cite{ji2024realtalk}. Each video is first segmented into temporally coherent clips using TransNetV2~\cite{soucek2024transnet}. For each clip, we separate the two-speaker audio streams with SAM Audio~\cite{shi2025sam}, obtaining synchronized audio $\mathbf{A}_{A}, \mathbf{A}_{B}$. On the visual side, we run MediaPipe~\cite{Lugaresi2019MediaPipeAF} to extract per-frame Blendshape parameters $\mathbf{b} \in \mathbb{R}^{52}$ and head rotation $\mathbf{R} \in \mathbb{R}^{3}$. The final dataset consists of synchronized tuples $\mathcal{D} = \{(\mathbf{A}_{A}, \mathbf{A}_{B}, \mathbf{B}_{A}, \mathbf{B}_{B}, \mathbf{R}_{A}, \mathbf{R}_{B})\}$.

\subsection{Human-to-Robot Facial Expression Mapping}
\label{sec: mapping}

\textbf{Mapping network.} At each frame $t$, the synthesis network predicts  $\hat{\mathbf{y}}_t=[\hat{\mathbf{b}}_t;\hat{\mathbf{r}}_t]$.
The head motion is realized by the neck module; given $\hat{\mathbf{r}}_t$, we compute the corresponding motor targets $\mathbf{u}^{\mathrm{head}}_t$ via an inverse-kinematics (IK) solver.
For the facial expressions, we exploit the semantic structure of the blendshape basis by partitioning the coefficient vector into region-specific subsets, denoted as $\hat{\mathbf{b}}_t=\{\hat{\mathbf{b}}^{(k)}_t\}_{k\in\mathcal{K}}$ with $\mathcal{K}=\{\mathrm{eyebrow},\mathrm{eyes},\mathrm{mouth},\mathrm{jaw}\}$.
For each region $k$, we train a lightweight inverse-mapping MLP that maps the local blendshape coefficients to the corresponding actuator command vector:
$\mathbf{u}^{(k)}_t = \pi_k\!\left(\hat{\mathbf{b}}^{(k)}_t\right)$.
Each regressor is trained with a standard Mean Squared Error objective over paired data $\{(\mathbf{b}^{(k)}_t,\mathbf{u}^{(k)}_t)\}$:
$\mathcal{L}^{(k)}_{\mathrm{map}} = \left\|\pi_k(\mathbf{b}^{(k)}_t)-\mathbf{u}^{(k)}_t\right\|_2^2.$
Then the final motor command $\mathbf{u}_t$ at frame $t$ is obtained (Fig.~\ref{fig:initeraction synthesis and mapping}(b)).

\textbf{Data collection.} For each physical head, we first calibrate the valid actuation ranges. We then generate $3000$ motor command vectors using Latin hypercube sampling within these bounds to ensure broad and uniform coverage of the feasible actuation space. 
For each sampled command $\mathbf{u}_i$, we capture a synchronized image $I_i$ of the face. From each image $I_i$, we estimate the corresponding facial parameters $(\mathbf{b}_i,\mathbf{r}_i)$ using MediaPipe~\cite{Lugaresi2019MediaPipeAF} and obtain the final supervision tuples $\{(\mathbf{u}_i,\mathbf{b}_i,\mathbf{r}_i)\}$ for training the inverse mapping networks.

%% file: sec/4_experiment.tex
\section{Experiment}
\label{sec:experiment}

\subsection{Experimental Setup}

\textbf{Hardware platform.}
Each robotic head provides 21 independently actuated facial degrees of freedom, driven by GuoHua 9g digital micro servos with a 180° operating range. To support higher-torque head orientation, the neck is actuated by three NEMA 17 stepper motors. All structural components were fabricated using Polylactic Acid (PLA) via FDM 3D printing. The silicone skin is magnetically attached to the underlying frame, enabling rapid installation and replacement.

\textbf{Onboard deployment.}
All inference runs on an NVIDIA Jetson AGX Orin (32 GB). For actuation, the Orin communicates via the I$^2$C protocol with cascaded PCA9685 PWM drivers, updating servo commands at 30\,Hz. The Orin also interfaces with an eye-mounted camera, a microphone for audio capture, and an onboard speaker for audio playback.

\subsection{Evaluation of the Automatic Design Flow}
\begin{table}[t]
\centering
\small
\caption{Comparison of optimization strategies for \textbf{automatic mechanism design}. We report success rate (SR), convergence time (CT), expressiveness score (Exp $ = \sum_k \alpha_k$), and intersection volume (IV). CT and Exp are reported \textbf{only} for successful runs; IV is $0$ for successful runs and otherwise computed from the final design at the maximum iteration.}
\renewcommand{\arraystretch}{1.0}
\resizebox{\linewidth}{!}{
\begin{tabular}{l | c c c c}
\toprule
Method & SR $(\%)$ $\uparrow$ & CT (s) $\downarrow$ & Exp $\uparrow$ & IV $(mm^3)$ $\downarrow$ \\
\midrule
Ours w/o outer loop & 20.0 & 25.7 & 8.69 & 2050.7 \\
Global Joint-Opt & 0.0 & -- & -- & 9690.5 \\
Local + Heuristic & 33.3 & 1098.2 & 8.57 & 1690.6 \\
\rowcolor{gray!20}
Ours & \textbf{66.7} & 591.3 & 8.53 &  \textbf{460.4}\\
\bottomrule
\end{tabular}
}
\label{tab:design_optimization_comparison}
\vspace{-0.5cm}
\end{table}

\textbf{Hierarchical optimization efficacy.}
We evaluate our hierarchical optimization framework on 15 heads with diverse facial geometries. We compare against three baselines: (i) \textit{Ours w/o outer loop} (local-only optimization), which performs module-level optimization but disables the outer-loop assembly refinement; (ii) \textit{Global Joint-Opt}, which jointly optimizes all design variables using a single objective that combines position-keeping and collision-penalty terms; and (iii) \textit{Local + Heuristic Repulsion}, which augments local optimization with a hand-crafted repulsion vector to resolve collisions during assembly (App.~\ref{sec: app-Optiomization baselines}). All methods are optimized using L-BFGS. Tab.~\ref{tab:design_optimization_comparison} summarizes the mean performance over 15 design instances, where all methods share the same initialization. \textit{Local-only} converges rapidly (25.7s) but fails in compact geometries, validating the indispensability of assembly-level refinement. The \textit{Global Joint-Opt} baseline fails in all cases: the coupled objective is highly non-convex, and collisions lack informative gradients for complex assemblies, leading to poor convergence. While \textit{Local + Heuristic} improves feasibility (5/15), its reliance on suboptimal centroid-based repulsion significantly prolongs convergence (1098.2s) and fails to resolve complex interlocks requiring amplitude scheduling. In contrast, our hierarchical framework achieves the highest success rate (10/15) and converges 1.9$\times$ faster than the heuristic baseline. Notably, the runtime difference between \textit{Local-only} and \textit{Ours} (25.7s vs. 591.3s) indicates that collision resolution dominates the computational cost; however, this investment is strictly justified by the substantial gain in realizability while maintaining comparable expressiveness.

\begin{figure}
    \centering
    \includegraphics[width=0.95\linewidth]{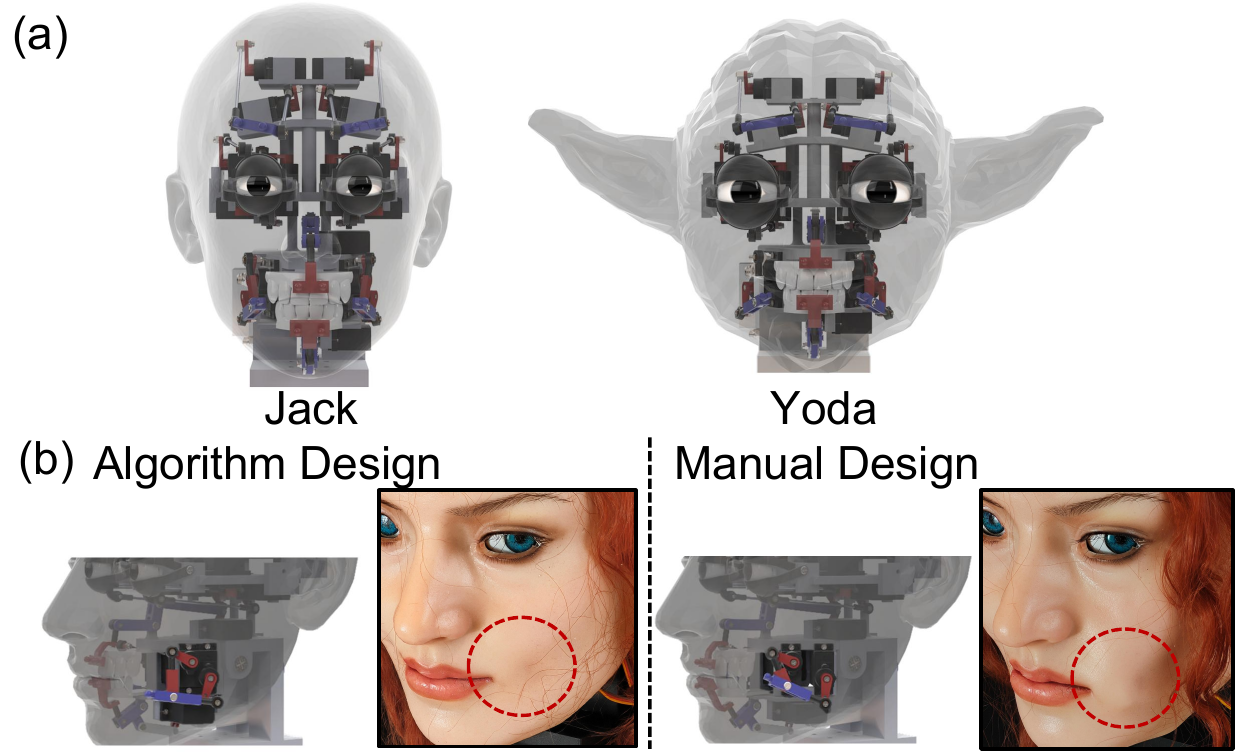}
    \caption{(a) \textbf{Algorithm design versatility}: CAD Assemblies for \textit{Jack} and \textit{Yoda}. (b) \textbf{Algorithm vs. Manual}: Comparison of mouth-corner mechanisms (red circles) on a compact head.}
    \label{fig:alogorithm comparison}
    \vspace{-0.1cm}
\end{figure}

Fig.~\ref{fig:alogorithm comparison}(a) shows qualitative results across diverse face geometries. We highlight two heads that deviate substantially from the template in both morphology and scale. \textbf{Left:} \textit{Jack} exhibits a sharper, more pointed chin than the template, which leaves less internal clearance and imposes stricter placement constraints for the mouth-corner and lower-lip mechanisms. \textbf{Right:} \textit{Yoda} has a markedly non-human facial structure; nevertheless, our pipeline successfully synthesizes a feasible internal mechanism layout. These examples demonstrate that our automatic design flow can handle large geometric variations and generalizes beyond human-like faces, indicating broad applicability and robustness.

\textbf{Comparative efficiency against manual design.} 
We further compare our automatic design flow with expert manual design on a compact head geometry (scaling factor $\beta=0.84$ relative to the template), which leaves limited internal volume for mechanisms. Two senior mechanical designers independently performed manual layouts and iterative adjustments, including rework triggered by interferences discovered during assembly. The average manual design time is 22.8\,h, substantially longer than our optimization runtime (11.7\,min). Fig.~\ref{fig:alogorithm comparison}(b) compares the resulting mouth-corner mechanism behaviors. Manual designs, driven by trial-and-error and implicit design heuristics, achieve weaker upward mouth-corner expressiveness and less skin-conforming motion trajectories, leading to reduced visual fidelity. In contrast, our method explicitly optimizes expressiveness under geometric constraints and produces a tighter, more consistent upward expressiveness.

\subsection{Evaluation of Conversational Motion Synthesis}
\textbf{Baselines and metrics.} We compare against three \textbf{speaker-only} methods (FaceFormer~\cite{fan2022faceformer}, EmoTalk~\cite{peng2023emotalk}, SelfTalk~\cite{peng2023selftalk}) and a speaker-listener baseline~\cite{peng2025dualtalk}. All methods are trained on our dataset. For the speaker, we evaluate lip synchronization (LVE), overall expression accuracy (MSE), head-pose dynamics (PDD), and motion diversity (SID). For the listener, we evaluate upper-face motion dynamics (FDD) and PDD.

\begin{table}[t]
\centering
\small
\caption{Comparison under \textbf{speaker} and \textbf{listener} modes on our dataset. ``RT'' means whether real-time use is supported.}
\renewcommand{\arraystretch}{1.0}
\resizebox{\linewidth}{!}{
\begin{tabular}{l | c c c c | c c | c}
\toprule
\multirow{2}{*}{Method} &
\multicolumn{4}{c|}{Speaker} &
\multicolumn{2}{c|}{Listener} & \multirow{2}{*}{RT}\\
\cmidrule(lr){2-5}\cmidrule(lr){6-7}
& LVE $\downarrow$ & MSE $\downarrow$ & PDD $\downarrow$ & SID $\uparrow$
& FDD $\downarrow$ & PDD $\downarrow$ \\
\midrule
FaceFormer  & 4.25 & 0.69 & 7.32 & 0.71 & -- & -- & \cmark\\
EmoTalk     & \textbf{2.98} & 0.68 & 6.88 & 3.03 & -- & -- &\cmark\\
SelfTalk    & 5.34 & 0.73 & 7.03 & 2.98 & -- & -- &\cmark\\
\midrule
DualTalk    & 3.79 & 0.57 & 8.74 & \textbf{3.36} & 29.77 & 9.45 &\xmark\\
\rowcolor{gray!20}
Ours    & 3.04 & \textbf{0.38} & \textbf{5.63} & 3.29 &\textbf{21.12} & \textbf{5.81} &\cmark\\
\bottomrule
\end{tabular}
}
\label{tab:dialogue_metrics}
\vspace{-0.3cm}
\end{table}

\textbf{Quantitative results.} As shown in Tab.~\ref{tab:dialogue_metrics}, our framework demonstrates superior performance across both conversational roles. In speaker mode, we achieve the lowest MSE and PDD, indicating that our generated expressions are both geometrically accurate and temporally stable while maintaining lip-sync quality. In listener mode, we outperform the dyadic baseline~\cite{peng2025dualtalk}. 
Crucially,~\cite{peng2025dualtalk} relies on the partner's future motion, restricting it to offline processing. In contrast, our model infers reactive behaviors solely from the two speakers' audio, thereby enabling real-time inference.

\subsection{Evaluation of Expression Mapping Results}

\begin{table}[t]
\centering
\small
\caption{Comparison for \textbf{mapping} grouped by facial representations. FPS is measured on Jetson AGX Orin.}
\renewcommand{\arraystretch}{1.0}
\resizebox{\linewidth}{!}{
\begin{tabular}{l | c c c | c | c >{\columncolor{gray!20}}c}
\toprule
\multirow{2}{*}{Metric} &
\multicolumn{3}{c|}{Landmarks} &
\multicolumn{1}{c|}{FLAME} &
\multicolumn{2}{c}{Blendshapes} \\
\cmidrule(lr){2-4}\cmidrule(lr){5-5}\cmidrule(lr){6-7}
& NNR & Norm.~\cite{chen2021smileeva} + MLP & Norm.~\cite{hu2024human} + MLP
&  MLP
&  MLP & Ours \\
\midrule
LSE-D $\downarrow$ & 15.59 & 14.52 & 13.74 & 13.06 & 12.68 & \textbf{10.61} \\
LSE-C $\uparrow$ & 0.34 & 1.94 & 2.39 & 2.42 & 2.58 & \textbf{3.19} \\
FPS $\uparrow$   & 6.36 & 6.31 & 6.19 & 2408.55 & \textbf{2478.26} & 2396.31 \\
\bottomrule
\end{tabular}
}
\vspace{-0.5cm}
\label{tab:mapping_baselines_metrics}
\end{table}

\textbf{Baselines and metrics.} We compare our semantic region-wise mapping against baselines spanning three common facial representations. \textbf{Landmarks}: (i) nearest-neighbor retrieval that matches each avatar landmark to the closest sample in the robot dataset, and two learning-based variants that reduce the avatar-robot gap via (ii)~\cite{chen2021smileeva} normalization and ~\cite{hu2024human} normalization followed by an MLP regressor. \textbf{FLAME}: an MLP that maps per-frame FLAME parameters to motor commands. \textbf{Blendshapes}: per-frame blendshape parameters to motor commands. All learning-based methods are trained with an MSE loss. We assess the robot's audio–lip synchronization using Lip Sync Error Distance (LSE-D) and Lip Sync Error Confidence (LSE-C)~\cite{prajwal2020lip}, as well as on-device FPS. Full-face expression fidelity is evaluated via our user study (Sec.~\ref{sec: user study}).

\textbf{Quantitative results.} Tab.~\ref{tab:mapping_baselines_metrics} shows comparisons across the baselines. Our semantic region-wise mapping achieves the best audio–lip synchronization while maintaining high throughput sufficient for real-time conversational interaction. Normalization-based landmark methods~\cite{zhang2025morpheus, chen2021smileeva, hu2024human} reduce the sim-to-real gap, but their performance is highly sensitive to the robot's facial geometry and the camera viewpoint. In addition, landmark pipelines introduce extra inference latency because they require rendering a 2D avatar frame and running landmark detection before generating motor commands. In contrast, FLAME and blendshape parameters provide identity-agnostic representations. However, a single per-frame MLP underperforms our approach because it must learn cross-region correspondences implicitly from data. By encoding semantic correspondences as an explicit prior, our network achieves superior synchronization with a smaller model.

\subsection{End-to-End Conversation Demonstration}

\begin{figure*}
    \centering
    \includegraphics[width=0.95\linewidth]{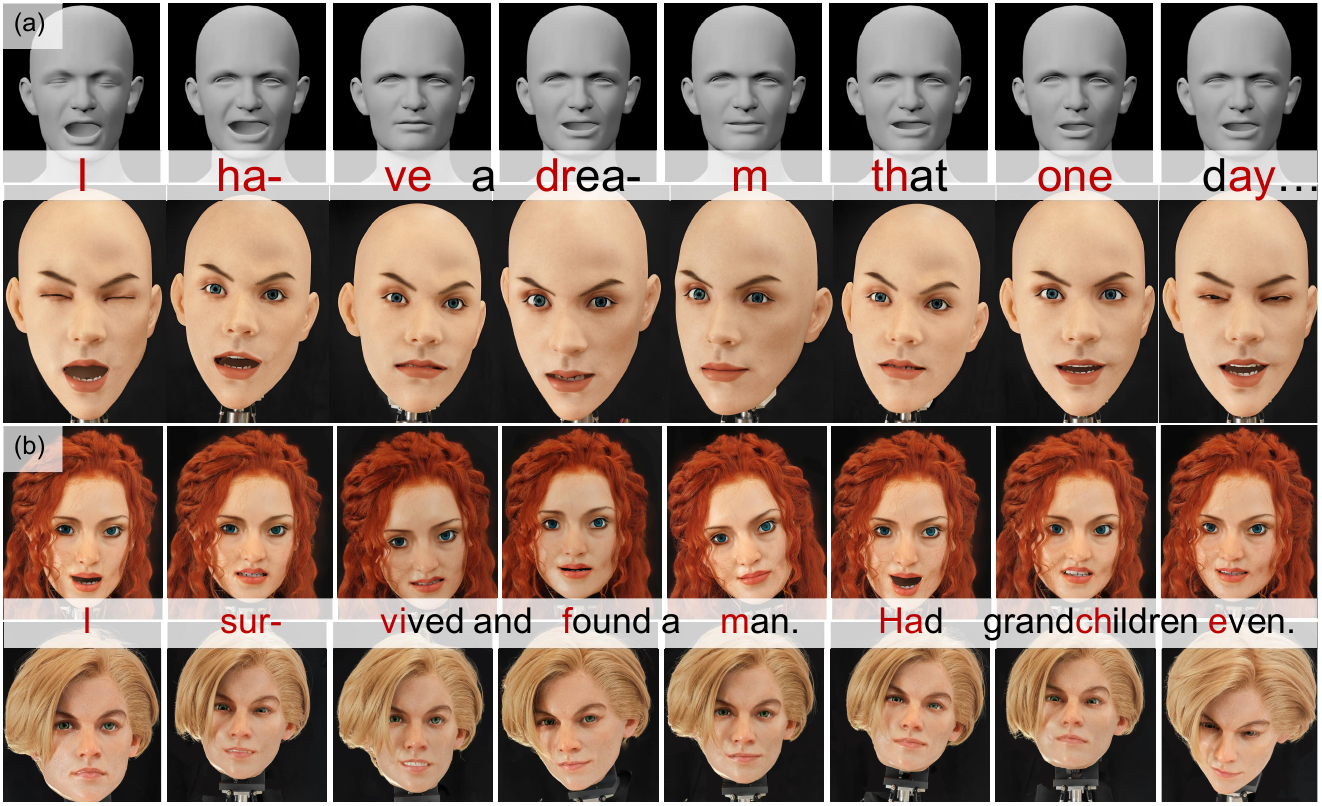}
    \caption{\textbf{End-to-end system visualization across interaction scenarios.} (a) \textbf{Human-Robot Interaction}: Real-time conversational gestures during a user-robot dialogue. The top row is the expression synthesis. (b) \textbf{Dyadic Role-Play (\textit{Titanic})}: Reenacting the scene dynamics with expressions corresponding to the dialogue between \textit{Rose} (speaker) and \textit{Jack} (listener).}
   \vspace{-0.3cm}
    \label{fig:visualization}
\end{figure*}

We show our end-to-end system in two interaction scenarios (Fig.~\ref{fig:visualization}). 
\textbf{Human-robot interaction.} We integrate the Doubao Voice Model as the conversational engine, enabling fluid, low-latency spoken dialogue with human users (Fig.~\ref{fig:visualization}(a)). 
\textbf{Dyadic robot-robot dialogue.} We stage a hypothetical reunion scene from \textit{Titanic}. Driven by input audio, the system generates high-fidelity lip synchronization for the speaker (\textit{Rose}) while synthesizing contextually congruent reactive expressions for the listener (\textit{Jack}) (Fig.~\ref{fig:visualization}(b)). 

\textbf{End-to-end build time breakdown.} Tab.~\ref{tab:time_breakdown} details the mean time required to realize a new character, from mechanism synthesis to a fully deployable robot. Our framework significantly compresses the mechanism design and mapping (Stages 1, 5, 6), which have traditionally served as the primary temporal bottlenecks in robotic customization. While the physical realization (Stages 2, 3, 4) currently occupies the majority of the timeline (measured with a single operator and machine), these processes are inherently parallelizable and can be further accelerated through scaled manufacturing resources.

\begin{table}[t]
\centering
\footnotesize
\renewcommand{\arraystretch}{0.5}
\caption{\textbf{End-to-end time breakdown.} 1: automatic design optimization; 2: 3D printing; 3: silicone face skin; 4: assembly; 5: mapping dataset collection; 6: mapping network training. The \textit{total} time assumes parallel execution of Stages 2 and 3.}
\renewcommand{\arraystretch}{1.0}
\resizebox{\linewidth}{!}{
\begin{tabular}{l | c c c c c c| c}
\toprule
Stage & 1 & 2 & 3 & 4 & 5 & 6 & Total \\
\midrule
Time (h) & 0.16 & 21.47 & 20.50 & 3.83 & 0.25 & 0.13 & 25.84 \\
\bottomrule
\end{tabular}
}
\vspace{-0.6cm}
\label{tab:time_breakdown}
\end{table}

\subsection{User study}
\label{sec: user study}
\begin{figure}
    \centering
    \includegraphics[width=0.95\linewidth]{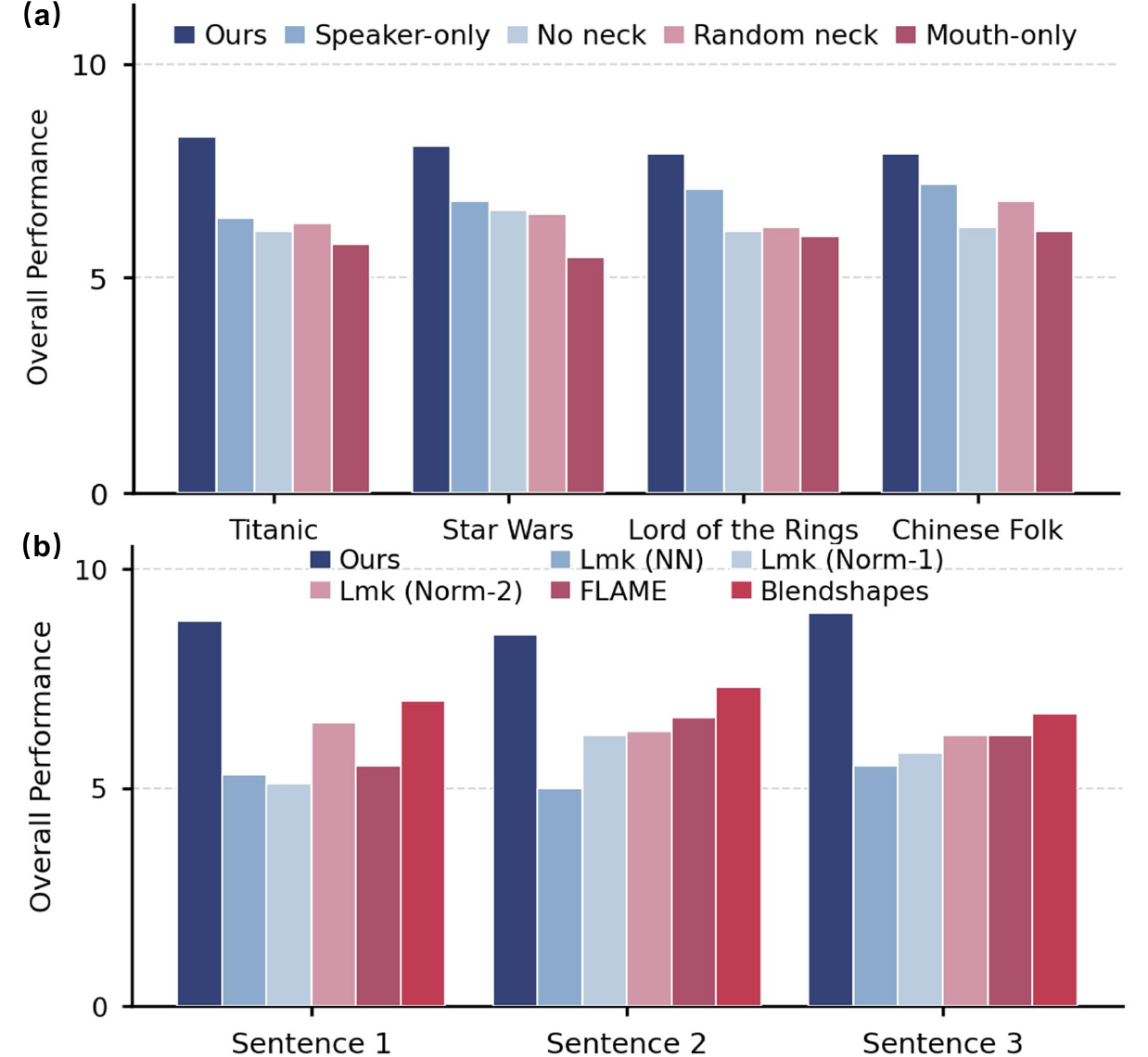}
    \caption{\textbf{User study results.} (a) Overall interaction performance on four excerpts. (b) Mapping quality on three sentences.}
    \label{fig:user study}
    \vspace{-0.5cm}
\end{figure}

To evaluate the perceived quality of our real-time interaction pipeline, we conducted a perceptual study using four representative dialogue excerpts. We recruited 100 participants. For each excerpt, participants were shown the reference clip together with five robot interaction variants: \textit{Ours}, \textit{speaker-only}, \textit{no neck motion}, \textit{random neck motion}, and \textit{mouth-only mapping}. Participants then rated the \textbf{overall performance} on a 10-point scale. Fig.~\ref{fig:user study}(a) shows that \textit{Ours} achieves the highest overall scores. The ablations suggest three consistent findings: (i) listener backchannel cues and subtle facial responses substantially improve perceived interaction quality, (ii) coordinated neck motion makes the delivery more natural than facial motion alone~\cite{zhang2025morpheus}, and (iii) realism cannot be explained by lip synchronization alone~\cite{hu2026learning, xu2026singingbot}, as full-face expressions provide essential cues beyond the mouth region.

We also evaluated mapping quality using three sentences. Fig.~\ref{fig:user study}(b) shows that our mapping produces more realistic and natural expressions, with a smaller gap to the reference motion(details in App.~\ref{sec: app-user study}).

%% file: sec/6_conclusion.tex
\section{Conclusion}
\label{sec:conclusion}
We present a unified framework that enables scalable synthesis for animatronic personalization. By coupling a collision-driven automated hardware design pipeline with a real-time conversational motion engine, we achieved rapid 2D-to-3D physical realization and immersive bidirectional interaction. This work lays the critical groundwork for a future where thousands of robots with distinct identities can be efficiently manufactured and integrated into human society.

%% file: sec/X_supp.tex
\clearpage

\appendix

\setcounter{figure}{0}
\setcounter{equation}{0}
\setcounter{table}{0}
\renewcommand{\thefigure}{A.\arabic{figure}}
\renewcommand{\theequation}{A.\arabic{equation}}
\renewcommand{\thetable}{A.\arabic{table}}
\label{sec:supp}

\subsection{Overview}
This appendix provides supplementary technical details, mathematical derivations, and extended experimental results to support the main findings of the paper. The content is organized as follows:

\begin{itemize} \item \textbf{App.~\ref{sec: app-extended related work}: Extended Related Work.} Provides a comprehensive literature review covering the evolution of mechanical face platforms, computational methods for automatic mechanism design, and state-of-the-art approaches in talking head generation and expression mapping.

\item \textbf{App.~\ref{sec: app-kinematics analysis}: Mechanism Kinematic Analysis.} Details the kinematic structures of individual facial modules (eyebrow, eyes, mouth, jaw, and neck). These mathematical models serve dual critical roles: they define the differentiable constraints for the Inner Loop Optimization during mechanism synthesis and provide the analytical inverse kinematics required for the motion mapping pipeline to drive the physical motors.

\item \textbf{App.~\ref{sec: app-optimization details}: Optimization Algorithm Details.} Elucidates the algorithmic implementation of the design automation framework, including geometric initialization procedures, design space definitions, and the mathematical formulation of comparative baselines. Additionally, it visualizes diverse synthesis results and analyzes failure cases.

\item \textbf{App.~\ref{sec: app-manufacture details}: Manufacture Details.} Specifications regarding the physical fabrication process, including 3D printing material, silicone skin fabrication, actuator selection, and noise characterization.

\item \textbf{App.~\ref{sec: app-talking head synthesis}: Talking Head Synthesis Details.} Describes the neural network architecture, training objective functions, the definitions of quantitative metrics (LVE, MSE, PDD, SID, FDD) used to evaluate the conversational motion synthesis framework, and our dataset details.

\item \textbf{App.~\ref{sec: app-mapping network}: Mapping Network Details.} Provides implementation details for the proposed semantic region-wise mapping network and defines the configurations for landmark-based, FLAME-based, and blendshape-based baselines used in the test. Furthermore, it presents an in-depth analysis of the performance trade-offs among these different mapping strategies.

\item \textbf{App.~\ref{sec: app-extended visualization}: Extended Visualization.} Presents qualitative results showcasing the system's capabilities across diverse interaction scenarios, including human-robot interaction, dyadic robot-robot dialogue, hybrid robot-avatar interaction.

\item \textbf{Sec.~\ref{sec: app-user study}: User Study Details.} Reports the methodology for the perceptual evaluation, including participant demographics, detailed experimental protocols for interaction and mapping quality studies, and statistical analysis of the subjective ratings.

\item \textbf{App.~\ref{sec: app-limitations and future work}: Limitations and Future Work.} Discusses current system constraints and outlines directions for future research, focusing on mechanical modularity, unified end-to-end learning, differentiable simulation, and full-body embodiment.

\end{itemize}

\subsection{Extended Related Work}
\label{sec: app-extended related work}
\subsubsection{Mechanical Face Platform}
Robotic face design has evolved significantly to enhance human-robot interaction, primarily driven by advancements in actuation modalities. Early research utilized flexible micro-actuators and pneumatic artificial muscles to replicate FACS-based motions, demonstrating the potential for mechanical emotional communication~\cite{kobayashi1993study,kobayashi2000study,kobayashi2003realization,hashimoto2006development,hashimoto2008dynamic,berns2006control,itoh2006mechanical,oh2006design}. To improve compliance and expression range, bio-inspired materials such as EAP actuators were introduced~\cite{hanson2002identity}, while rope- and tendon-driven mechanisms were explored to achieve more compact and human-like deformations~\cite{weiguo2004development,habib2014learning,fortunati2021rise,li2024driving}.
Parallel efforts in system integration saw the emergence of full-body humanoids, such as HRP-4C~\cite{kaneko2009cybernetic,nakaoka2009creating}, the EveR series~\cite{lee2008development,ahn2012designing}, and Janet and Thomas~\cite{lin2009realization}, as well as android heads like PKD, Sophia, and Erica~\cite{habib2014learning,fortunati2021rise,glas2016erica}, which embedded tens of facial DoFs into anthropomorphic forms. Developmental and socially assistive platforms, including Cog, Flobi, Affetto, and minimalist muscle-like designs, further investigated the interplay between actuation density, sensing, and social engagement~\cite{brooks1998cog,lutkebohle2010bielefeld,ishihara2011realistic,allison2008design}.

More recently, the field has shifted towards densifying facial DoFs and refining transmission architectures to achieve hyper-realistic interaction. Contemporary platforms such as Abel~\cite{cominelli2021abel}, Ameca~\cite{nieto2025robot}, Eva~\cite{faraj2021facially,chen2021smileeva}, EMO~\cite{hu2024human}, and Nikola~\cite{yang2022optimizing,yang2025hapi} leverage advanced servo-linkage and pneumatic systems for real-time expressiveness. This trend continues with latest designs exploring diverse points in the design space: the Xiao Yao head~\cite{sheng2025review} and HHU robot~\cite{liu2022real} refine kinematic structures, while Yu-F01~\cite{fan2025soft} targets low-cost soft-skin solutions. Similarly, Rena~\cite{zhu2025awakening} emphasizes robust mechanisms for long-term interaction, and Morpheus~\cite{zhang2025morpheus} introduces a 33-DoF hybrid actuation scheme to maximize the span of realizable expressions. 
However, a critical limitation persists across these state-of-the-art systems: each is meticulously engineered for a single target face with a fixed mechanical layout. Consequently, adapting these sophisticated architectures to a new 3D face shape typically necessitates substantial manual redesign. In contrast, our work presents a scalable face robot platform designed to transfer the same high-fidelity actuation scheme to diverse 3D geometries with minimal engineering effort.

\subsubsection{Automatic Robot Design}

Computer graphics research has extensively explored synthesizing mechanisms from motion specifications, ranging from planar linkages~\cite{coros2013computational, bacher2015linkedit} and linkage-based characters~\cite{thomaszewski2014computational} to automata derived from motion capture~\cite{ceylan2013designing, zhu2012motion}. Complementary tools support rapid fabrication~\cite{megaro2014chacra, cali20123d} and the retargeting of mechanism templates to arbitrary 3D geometries~\cite{zhang2017functionality}.
However, these methods prioritize kinematic feasibility over the physical constraints essential for real-world robotics. Unlike our framework, they typically overlook internal interference checks required for operating within highly constrained volumes.

In robotics, optimization extends to dynamics and actuation, with task-based methods widely used to tune cable-driven mechanisms for workspace and stiffness requirements~\cite{li2017task, kawaharazuka2023design, islam2024task, penta2016analysis}. 
At the system level, morphology-control co-optimization employs differentiable, RL, and graph-based frameworks~\cite{xu2021end, ha2019reinforcement, schaff2019jointly, he2024morph, wang2019neural, zhao2020robogrammar, kawaharazuka2024robot, li2023modular, sun2026knowledge} across domains ranging from locomotion~\cite{geilinger2018skaterbots, ha2016taskleg} and soft robotics~\cite{lipson2000automatic, hiller2011automatic} to manipulation~\cite{mannam2024design, gilday2025embodied, gilday2024exploiting, yi2025codesign, kodnongbua2022computationalgripper} and multicopters~\cite{du2016computational, xu2019learning}. 
However, such search- / learning-based strategies typically yield coarse topologies. They lack the parametric precision required to embed complex linkages into thin-walled facial shells without compromising the delicate expressiveness of the skin.

Specific to the domain of facial robotics, seminal works have addressed the physical coupling between rigid actuators and deformable skin to achieve realistic expression cloning~\cite{bickel2012physical, huber2021designing}. However, these methods are primarily restricted to optimizing actuator placement or skin thickness, tend to treat the kinematic structure as a predefined prior.
Instead, We propose a framework for automatically retargeting high-DOF facial mechanisms that unifies kinematic synthesis with engineering validation. By incorporating trajectory fitting, spatial expressiveness, interference constraints into a closed optimization loop, we guarantee designs that are both expressive and manufacturable.

\subsubsection{Talking Head Generation and Expression Mapping}

Although 2D audio-driven talking face generation achieves impressive visual fidelity~\cite{prajwal2020lip, zhang2023sadtalker, kong2024hunyuanvideo, peng2025omnisync, lin2025omnihuman}, these methods typically suffer from high generation latency~\cite{zhou2025interactive,zhu2025infp, xu2026singingbot, hu2026learning} and fail to provide robust mapping perspectives for multi-person dialogues~\cite{zhang2025shouldershot}.
Consequently, 3D talking head synthesis~\cite{cudeiro2019capture, richard2021meshtalk,chu2025artalk} is preferred for robotic actuation~\cite{zhang2025morpheus}. Foundational sequence-to-sequence models established deterministic lip-sync capabilities~\cite{fan2022faceformer, xing2023codetalker}, while subsequent research integrated emotional expressiveness~\cite{peng2023emotalk, danvevcek2023emotional, liu2024emoface} and enhanced cross-identity generalization through implicit or unified representations~\cite{lu2025lsf, fan2024unitalker, chen2025cafe}.
Crucially for conversational robots, recent focus has shifted from monologue to dynamic interaction. While some works prioritize synchronization quality~\cite{peng2023selftalk, peng2024synctalk, peng2025synctalk++}, emerging dual-turn synthesis frameworks~\cite{peng2025dualtalk,ng2024audio, tran2024dim} explicitly model the reciprocal dynamics of interactions. This shift enables the generation of reactive listening behaviors, a prerequisite for sustaining realistic multi-turn dialogues.

To map synthesized motions to physical robots, standard approaches typically rely on geometric correspondence or linear blendshape mapping~\cite{chen2021smileeva, fan2025soft, wu2024retargeting, heisler2025iphone, zhu2025awakening}.
Recent efforts advance this by employing semantic representations (e.g., AUs)~\cite{liu2024unlocking}, data-driven learning frameworks~\cite{li2025x2c, zhang2025exface, zhang2025fabg}, and optimization techniques~\cite{yang2025hapi, li2024driving} to bridge the non-linear gap between human expressions and robotic actuation. Unlike prior works limited to single-frame expression mapping or single-speaker generation, we jointly model dual-turn synthesis with physical mapping to realize robust multi-turn conversational interaction on mechanical robots.

\subsection{Mechanism Kinematic Analysis}
\label{sec: app-kinematics analysis}

\begin{figure}
    \centering
    \includegraphics[width=1.0\linewidth]{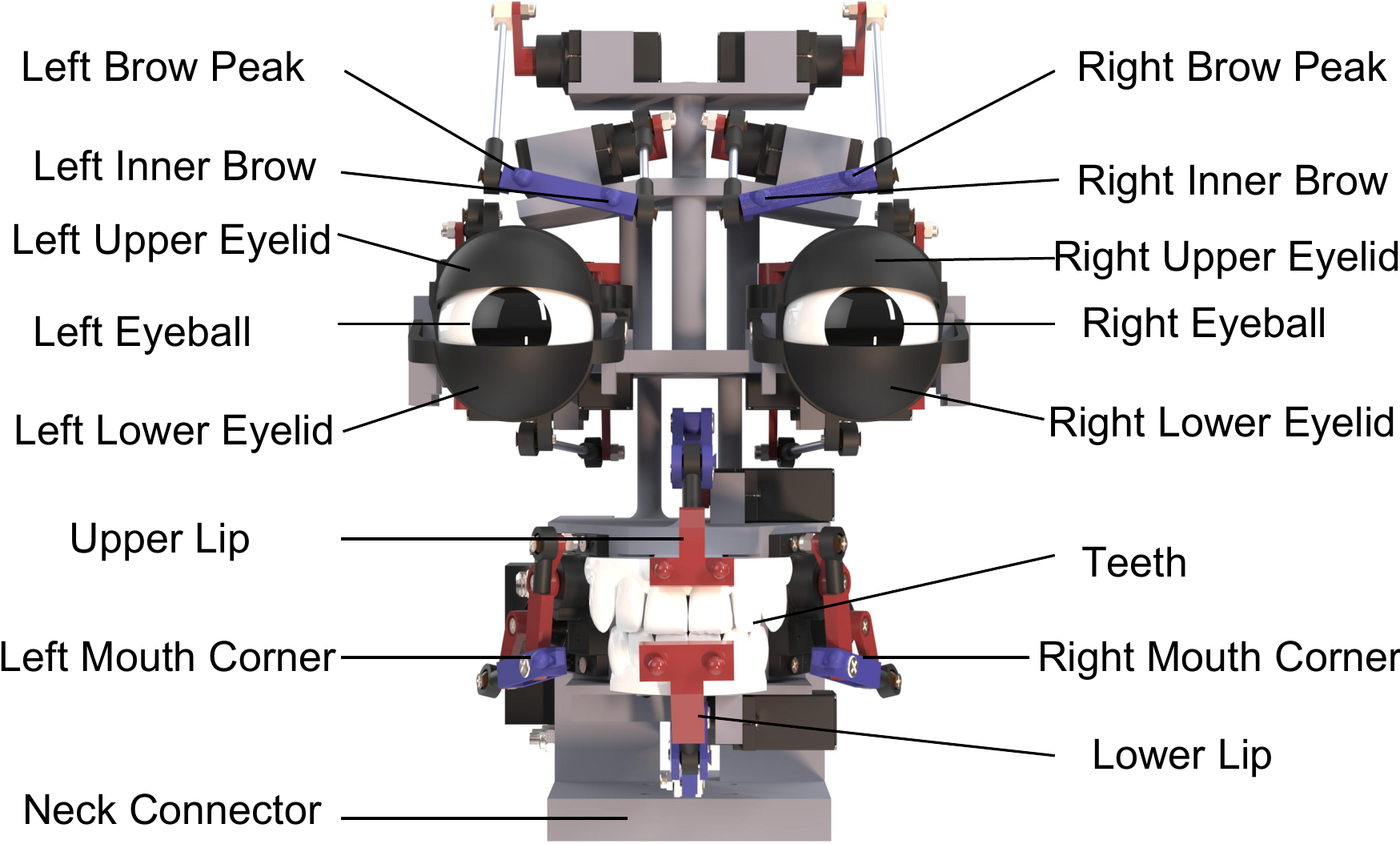}
    \caption{Detailed mechanical of the facial robot template.}
    \label{fig:template details}
\end{figure}

In this section, we provide detailed kinematic derivations for the constituent facial modules, including the \textbf{eyebrow}, \textbf{eye}, \textbf{mouth}, \textbf{jaw}, and \textbf{neck} mechanisms (Fig.~\ref{fig:template details}). We rigorously define the mathematical formulations of the specific kinematic parameters that serve as the objective functions within the \textbf{Inner Loop Optimization} of our hierarchical design framework. Finally, to benchmark architectural extensibility, we apply the identical optimization protocol to a state-of-the-art open-source baseline. This comparative analysis explicitly highlights the superior scalability of our modular design.

\begin{figure*}
    \centering
    \includegraphics[width=1.0\linewidth]{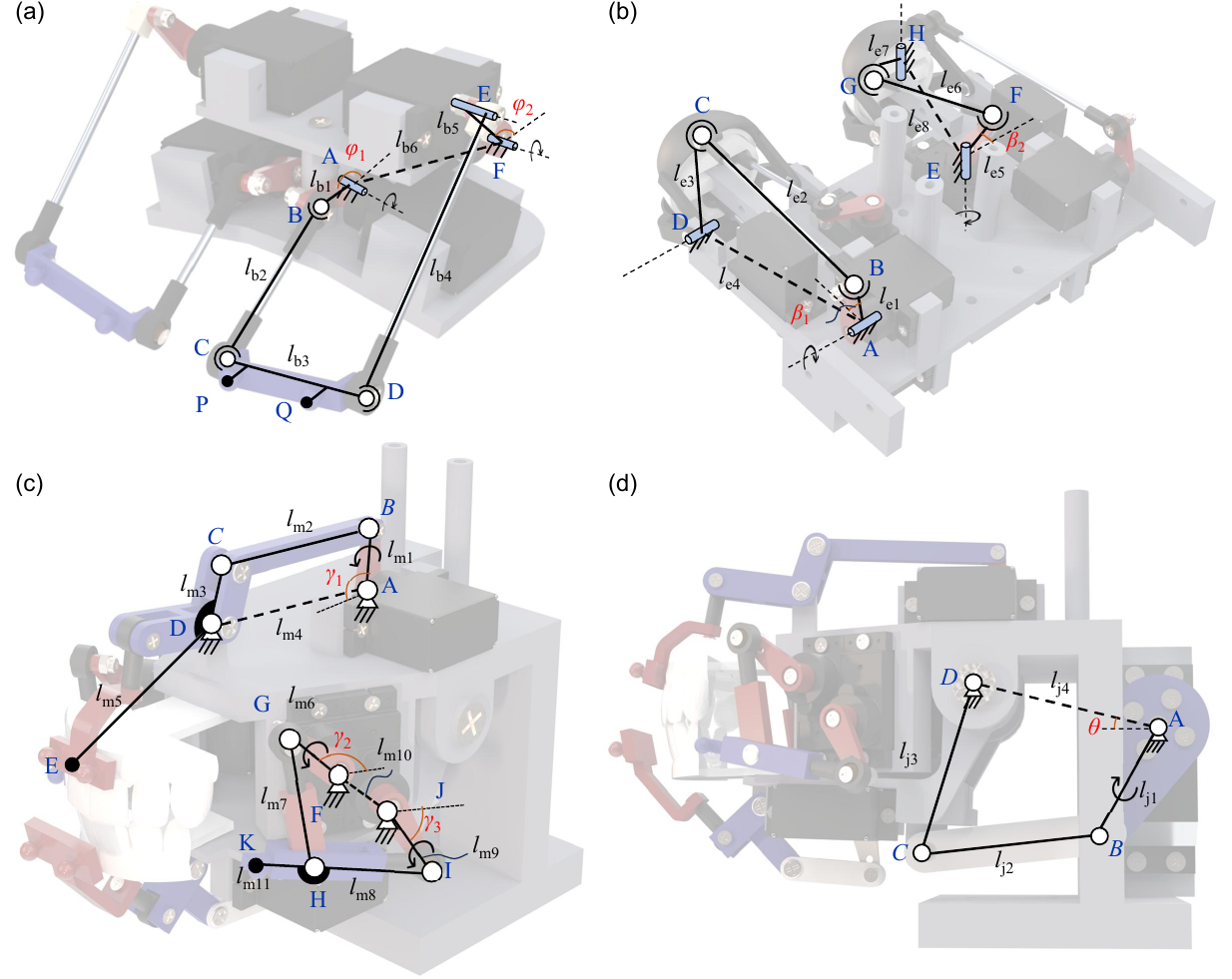}
    \caption{\textbf{Kinematic schematics of the facial actuation modules.} The CAD renderings are overlaid with simplified linkage diagrams, where labeled nodes denote joints and $l_i$
 denote link lengths. (a) \textbf{Eyebrow module}: a symmetric six-bar linkage (left/right share the same design). (b) \textbf{Eyes module}: two four-bar linkages, with the left linkage actuating eyelid blinking and the right linkage actuating eyeball rotation. (c) \textbf{Mouth module}: a four-bar linkage for lip motion and a five-bar linkage for mouth-corner actuation. (d) \textbf{Jaw module}: a four-bar linkage that controls jaw opening/closing.}
    \label{fig:app-mechanism details}
\end{figure*}

\subsubsection{Eyebrow module}
Fig.~\ref{fig:app-mechanism details}(a) shows the eyebrow mechanism, implemented as a spatial six-bar linkage with left/right symmetry. From the ground to the output member, the closed chain can be viewed as a kinematic sequence $R$--$S$--$S$--$S$--$R$--$R$, where $R$ denotes a revolute joint and $S$ denotes a spherical joint.
We define a global frame $\{W\}$.
Two ground pivots $A$ and $E$ are fixed in $\{W\}$ and actuated by revolute angles $\varphi_1$ and $\varphi_2$, respectively.
The primary output member is the rigid bar between spherical joints $C$ and $D$ (link $l_{b3}$), on which two task points $P$ and $Q$ are rigidly attached.
To achieve precise spatial guidance, we impose an external planar constraint: both $P$ and $Q$ must lie on a predefined plane $\mathcal{P}$ in the global frame,
\begin{equation}
\mathcal{P} := \{\mathbf{x}\in\mathbb{R}^3 \mid \mathbf{n}^\top \mathbf{x} + d = 0\},
\label{eq:plane_def}
\end{equation}
where $\mathbf{n}$ is the unit normal and $d$ is the signed offset. This constraint couples the spatial motion of the linkage with planar trajectory control of the skin-contact points.

\textbf{Mobility and constraint evaluation.} 
We evaluate the mobility using the modified Grübler--Kutzbach criterion for spatial mechanisms:
\begin{equation}
M_{\mathrm{raw}} = 6(n-1-j) + \sum_{i=1}^{j} f_i,
\label{eq:gk}
\end{equation}
where $n$ is the number of links (including ground), $j$ is the number of joints, and $f_i$ is the DOF of the $i$-th joint.
Here, $n=6$ and the loop contains $j=6$ joints comprising three revolute joints ($f=1$) and three spherical joints ($f=3$), yielding
\begin{equation}
M_{\mathrm{raw}} = 6(6-1-6) + (3\cdot 1 + 3\cdot 3) = 6.
\label{eq:Mraw}
\end{equation}
Two intermediate connecting rods are modeled as axially symmetric links connected by spherical joints; their twist about the longitudinal axis does not affect the mechanism geometry and thus constitutes two idle (passive) DOFs.
Removing these yields an effective spatial mobility of $M_{\mathrm{eff,spatial}}=4$.
Finally, constraining both $P$ and $Q$ to plane $\mathcal{P}$ introduces two independent scalar constraints, resulting in
\begin{equation}
M_{\mathrm{eff}} = 4 - 2 = 2,
\label{eq:Meff}
\end{equation}
i.e., the mechanism is a 2-DOF system fully determined by the two inputs $\mathbf{q}=[\varphi_1,\varphi_2]^\top$.

\textbf{Forward kinematics formulation.}
Let $\mathbf{a},\mathbf{e}\in\mathbb{R}^3$ denote the fixed positions of pivots $A$ and $E$ in $\{W\}$.
The actuated revolute joints generate the positions of adjacent joints $B$ and $F$ via
\begin{equation}
\mathbf{b}(\varphi_1) = \mathbf{a} + \mathbf{R}_A(\varphi_1)\,\mathbf{r}_{AB},
\label{eq:b_phi1}
\end{equation}
\begin{equation}
\mathbf{f}(\varphi_2) = \mathbf{e} + \mathbf{R}_E(\varphi_2)\,\mathbf{r}_{EF},
\label{eq:f_phi2}
\end{equation}
where $\mathbf{r}_{AB}$ and $\mathbf{r}_{EF}$ are constant link vectors defined in local frames, and $\mathbf{R}_A(\cdot),\mathbf{R}_E(\cdot)$ are axis-angle rotations about the corresponding revolute axes.

Given $\mathbf{b}(\varphi_1)$ and $\mathbf{f}(\varphi_2)$, we choose the unknowns as the positions of spherical joints $\mathbf{c},\mathbf{d}\in\mathbb{R}^3$.
The loop closure through spherical joints yields the distance invariants
\begin{equation}
\Phi_1(\mathbf{c}) = \|\mathbf{c}-\mathbf{b}(\varphi_1)\|_2^2 - l_{b2}^2 = 0,
\label{eq:dist_bc}
\end{equation}
\begin{equation}
\Phi_2(\mathbf{c},\mathbf{d}) = \|\mathbf{d}-\mathbf{c}\|_2^2 - l_{b3}^2 = 0,
\label{eq:dist_cd}
\end{equation}
\begin{equation}
\Phi_3(\mathbf{d}) = \|\mathbf{d}-\mathbf{f}(\varphi_2)\|_2^2 - l_{b4}^2 = 0.
\label{eq:dist_df}
\end{equation}
Since joint $F$ is revolute (not spherical), an additional axis-consistency constraint is required.
Let $\hat{\mathbf{s}}_{F}(\varphi_2)$ be the unit direction of the revolute axis at $F$ expressed in the world frame, and let $h_F$ be a constant determined by the local placement of joint $D$ in the link attached to $F$.
The rigid attachment of the $DF$ link to the revolute joint implies
\begin{equation}
\Phi_4(\mathbf{d}) = \hat{\mathbf{s}}_{F}(\varphi_2)^\top\big(\mathbf{d}-\mathbf{f}(\varphi_2)\big) - h_F = 0.
\label{eq:rev_axis_constraint}
\end{equation}

We model $P$ and $Q$ as fixed points on the output bar $CD$.
With constant scalars $\eta_P,\eta_Q\in\mathbb{R}$,
\begin{equation}
\mathbf{p}(\mathbf{c},\mathbf{d}) = \mathbf{c} + \eta_P(\mathbf{d}-\mathbf{c}), 
\qquad
\mathbf{q}(\mathbf{c},\mathbf{d}) = \mathbf{c} + \eta_Q(\mathbf{d}-\mathbf{c}).
\label{eq:pq_on_cd}
\end{equation}
The planar constraint \eqref{eq:plane_def} yields
\begin{equation}
\Psi_1(\mathbf{c},\mathbf{d}) = \mathbf{n}^\top \mathbf{p}(\mathbf{c},\mathbf{d}) + d = 0,
\label{eq:plane_p}
\end{equation}
\begin{equation}
\Psi_2(\mathbf{c},\mathbf{d}) = \mathbf{n}^\top \mathbf{q}(\mathbf{c},\mathbf{d}) + d = 0.
\label{eq:plane_q}
\end{equation}

Collecting \eqref{eq:dist_bc}--\eqref{eq:plane_q}, we obtain a nonlinear algebraic system
\begin{equation}
\mathbf{F}(\mathbf{x};\mathbf{q})=\mathbf{0}, \qquad
\mathbf{x}:=\begin{bmatrix}\mathbf{c}^\top & \mathbf{d}^\top\end{bmatrix}^\top,\;\;
\mathbf{q}:=\begin{bmatrix}\varphi_1 & \varphi_2\end{bmatrix}^\top,
\label{eq:F_system}
\end{equation}
which uniquely determines $\mathbf{c},\mathbf{d}$ (and thus $P,Q$) for a given input $\mathbf{q}$ under the 2-DOF setting.

Because \eqref{eq:F_system} is nonlinear, we solve it via Newton--Raphson iterations:
\begin{equation}
\mathbf{x}^{(t+1)} = \mathbf{x}^{(t)} - \mathbf{J}(\mathbf{x}^{(t)};\mathbf{q})^{-1}\,\mathbf{F}(\mathbf{x}^{(t)};\mathbf{q}),
\label{eq:newton}
\end{equation}
where the configuration Jacobian is
\begin{equation}
\mathbf{J}(\mathbf{x};\mathbf{q}) = \frac{\partial \mathbf{F}}{\partial \mathbf{x}}(\mathbf{x};\mathbf{q}).
\label{eq:jacobian}
\end{equation}
In practice, we initialize $\mathbf{x}$ using the converged solution from the previous timestep (continuation over $\mathbf{q}$), which yields rapid and robust convergence.
The solved trajectories $\mathbf{p}(\mathbf{c},\mathbf{d})$ and $\mathbf{q}(\mathbf{c},\mathbf{d})$ enable workspace boundary estimation and singularity detection by monitoring $\det(\mathbf{J})$ or the minimum singular value of $\mathbf{J}$.

\subsubsection{Eyes module}
As shown in Fig.~\ref{fig:app-mechanism details}(b), each eye integrates four four-bar linkages: two four-bar mechanisms drive the upper and lower eyelids for blinking, and two additional four-bar mechanisms actuate the eyeball for left–right (yaw) motion. Altogether, the binocular eye module comprises eight four-bar linkages, enabling independent yet coordinated control of eyelid closure and gaze direction.

\subsubsection{Mouth module}
The mouth module comprises four subcomponents: the upper lip, lower lip, left mouth corner, and right mouth corner (Fig.~\ref{fig:app-mechanism details}(c)). The upper and lower lips are each actuated by a symmetric four-bar linkage; the output attachment point $E$ couples to the skin to generate controlled upward/downward lip motion. The mouth corners are driven by five-bar linkages, where the skin attachment point 
$K$ enables a richer corner trajectory, supporting upward raising, lateral translation, and downward pulling.

\textbf{Upper/Lower lip.} We model the upper/lower-lip actuation as a planar four-bar linkage with link lengths:
input crank $l_{m1}=|AB|$, coupler $l_{m2}=|BC|$, follower $l_{m3}=|CD|$, and ground link $l_{m4}=|AD|$.
In the global frame, the fixed pivots are $A=(x_A^m,y_A^m)$ and $D=(x_D^m,y_D^m)$, and the input angle $\gamma$ is measured from the global $x$-axis.
The moving joint $B$ is
\begin{equation}
x_B^m = x_A^m + l_{m1}\cos\gamma,\qquad
y_B^m = y_A^m + l_{m1}\sin\gamma.
\label{eq:mouth_B}
\end{equation}

The joint $C=(x_C^m,y_C^m)$ is determined by the intersection of two circles centered at $B$ and $D$:
\begin{equation}
(x_C^m-x_B^m)^2+(y_C^m-y_B^m)^2 = l_{m2}^2,
\label{eq: mouth_C_circles 1}
\end{equation}
\begin{equation}
(x_C^m-x_D^m)^2+(y_C^m-y_D^m)^2 = l_{m3}^2.
\label{eq:mouth_C_circles 2}
\end{equation}
Let $\mathbf{b}=[x_B^m,y_B^m]^\top$, $\mathbf{d}=[x_D^m,y_D^m]^\top$, and $\mathbf{r}=\mathbf{d}-\mathbf{b}$ with $s=\|\mathbf{r}\|_2$.
A closed-form solution of \eqref{eq: mouth_C_circles 1}\eqref{eq:mouth_C_circles 2} is given by
\begin{equation}
a=\frac{l_{m2}^2-l_{m3}^2+s^2}{2s},\qquad
h=\sqrt{\max(l_{m2}^2-a^2,\,0)},
\label{eq:mouth_ah}
\end{equation}
\begin{equation}
\mathbf{c}=\mathbf{b}+\frac{a}{s}\mathbf{r}\ \pm\ \frac{h}{s}\mathbf{r}_\perp,\qquad
\mathbf{r}_\perp=[-r_y,\ r_x]^\top,
\label{eq:mouth_C_closedform}
\end{equation}
where the $\pm$ corresponds to the two assembly modes (``elbow-up'' / ``elbow-down'').
In practice, we select the physically valid branch by continuity from the previous timestep and by joint-limit constraints.

Once $C$ is obtained, the follower orientation is
\begin{equation}
\theta_{CD}=\mathrm{atan2}(y_C^m-y_D^m,\ x_C^m-x_D^m).
\label{eq:mouth_thetaCD}
\end{equation}
To express an arbitrary output/attachment point $E$ rigidly fixed on the follower, we specify its polar parameters relative to the pivot $D$: a constant radius $l_{m5}=|DE|$ and a constant offset angle $\delta$ with respect to link $CD$.
The global coordinates of $E$ are then
\begin{equation}
x_E^m = x_D^m + l_{m5}\cos(\theta_{CD}+\delta),\qquad
y_E^m = y_D^m + l_{m5}\sin(\theta_{CD}+\delta).
\label{eq:mouth_E}
\end{equation}
This forward-kinematics mapping provides a concise way to propagate the input crank motion to lip skin-contact points and to analyze the resulting output trajectories for mechanism synthesis and optimization.

\textbf{Mouth corner.} 
We model the mouth-corner actuation as a planar five-bar linkage consisting of two actuated cranks $GF$ and $IJ$, two passive links $GH$ and $HI$, and a fixed base $FJ$.
The fixed pivots are $F=(x_F^m,y_F^m)$ and $J=(x_J^m,y_J^m)$ in the global frame.
The base length and its installation angle are
\begin{equation}
l_{m10} = \sqrt{(x_F^m-x_J^m)^2+(y_F^m-y_J^m)^2},
\label{eq:mouth5_base_len}
\end{equation}
\begin{equation}
\Gamma = \mathrm{atan2}\!\left(y_J^m-y_F^m,\; x_J^m-x_F^m\right).
\label{eq:mouth5_base_angle}
\end{equation}
The mechanism has two independent inputs, denoted by $\gamma_2$ and $\gamma_3$, which specify the driving-link angles measured relative to the base direction.
Let $\mathbf{q}=[\gamma_2,\gamma_3]^\top$.

Given $\mathbf{q}$, the global positions of the driving joints $G$ and $I$ are
\begin{equation}
\mathbf{r}_G =
\begin{bmatrix}
x_G^m\\y_G^m
\end{bmatrix}
=
\begin{bmatrix}
x_F^m\\y_F^m
\end{bmatrix}
+
l_{m6}
\begin{bmatrix}
\cos(\Gamma+\gamma_2)\\
\sin(\Gamma+\gamma_2)
\end{bmatrix},
\label{eq:mouth5_G}
\end{equation}
\begin{equation}
\mathbf{r}_I =
\begin{bmatrix}
x_I^m\\y_I^m
\end{bmatrix}
=
\begin{bmatrix}
x_J^m\\y_J^m
\end{bmatrix}
+
l_{m9}
\begin{bmatrix}
\cos(\Gamma+\gamma_3)\\
\sin(\Gamma+\gamma_3)
\end{bmatrix}.
\label{eq:mouth5_I}
\end{equation}

The floating joint $H=(x_H^m,y_H^m)$ satisfies the distance invariants
\begin{equation}
\|\mathbf{r}_H-\mathbf{r}_G\|_2 = l_{m7},\qquad
\|\mathbf{r}_H-\mathbf{r}_I\|_2 = l_{m8}.
\label{eq:mouth5_H_circles}
\end{equation}
A feasible assembly requires
\begin{equation}
|l_{m7}-l_{m8}| \le \|\mathbf{r}_I-\mathbf{r}_G\|_2 \le l_{m7}+l_{m8}.
\label{eq:mouth5_existence}
\end{equation}
Let $\mathbf{r}=\mathbf{r}_I-\mathbf{r}_G$ and $s=\|\mathbf{r}\|_2$.
A closed-form solution of \eqref{eq:mouth5_H_circles} is
\begin{equation}
a=\frac{l_{m7}^2-l_{m8}^2+s^2}{2s},\qquad
h=\sqrt{\max(l_{m7}^2-a^2,\,0)},
\label{eq:mouth5_ah}
\end{equation}
\begin{equation}
\mathbf{r}_H=\mathbf{r}_G+\frac{a}{s}\mathbf{r}\ \pm\ \frac{h}{s}\mathbf{r}_\perp,\qquad
\mathbf{r}_\perp=\begin{bmatrix}-r_y\\ r_x\end{bmatrix},
\label{eq:mouth5_H_closed}
\end{equation}
where the $\pm$ corresponds to the two assembly modes; we select the physically valid branch by continuity and joint-limit constraints.

The skin attachment point $K$ is constrained to lie on the extension of link $HI$ with a constant offset $l_{m11}$ from joint $H$, i.e., $HK=l_{m8}+l_{m11}$ and $H$, $I$, $K$ are collinear.
Thus,
\begin{equation}
\mathbf{r}_K=\mathbf{r}_H+\frac{l_{m8}+l_{m11}}{l_{m8}}\left(\mathbf{r}_I-\mathbf{r}_H\right).
\label{eq:mouth5_K}
\end{equation}
This mapping shows that the reachable set of $K$ depends on the link-length ratios, admissible input ranges $(\gamma_2,\gamma_3)$, and the global installation angle $\Gamma$.
By scanning the feasible domain of $\mathbf{q}$ subject to \eqref{eq:mouth5_existence} and joint limits, the workspace boundary of $K$ can be computed for subsequent trajectory fitting and optimization.

\subsubsection{Jaw module}
As illustrated in Fig.~\ref{fig:app-mechanism details}(d), jaw opening/closing is actuated by a compact four-bar linkage. Points $A$ and $D$ are fixed ground pivots, while the remaining links transmit the actuator motion to the jaw output.

\subsubsection{Neck module}

\begin{figure}
    \centering
    \includegraphics[width=0.85\linewidth]{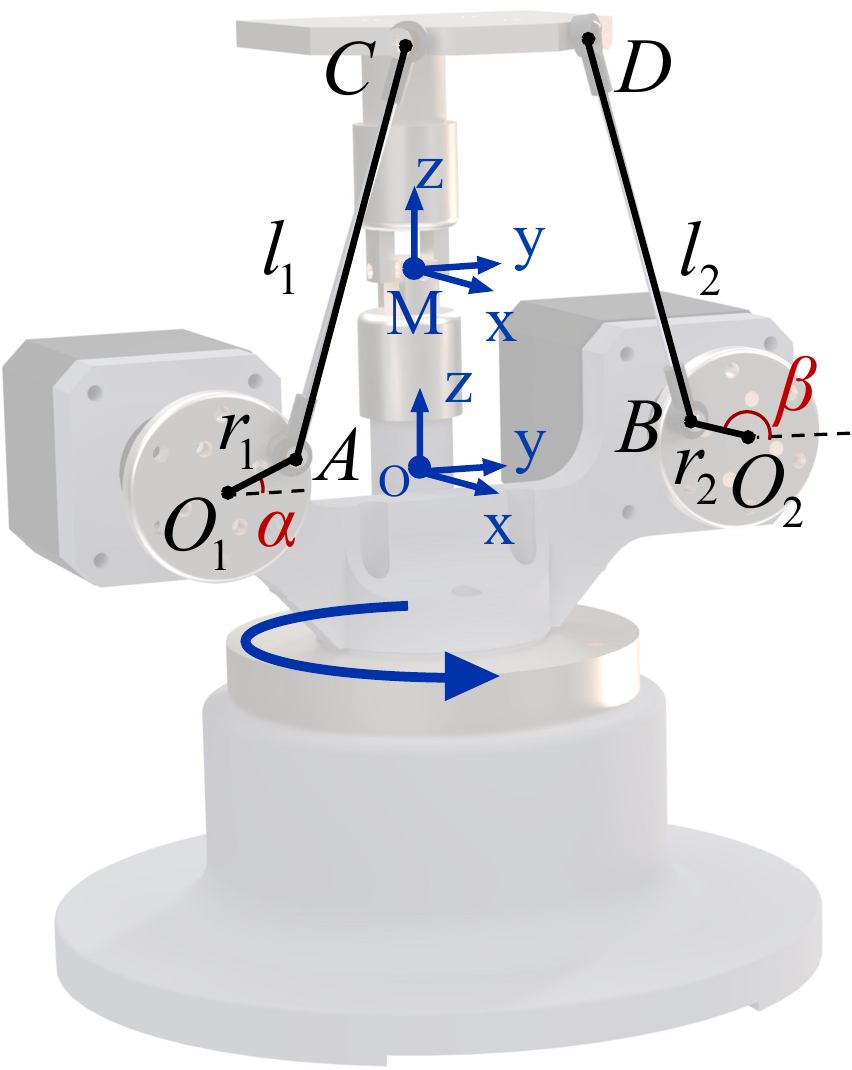}
    \caption{Neck module mechanism.}
    \label{fig:app-neck kinematics}
\end{figure}
This section derives the closed-form \textbf{inverse kinematics} (IK) for the neck mechanism in Fig.~\ref{fig:app-neck kinematics}. 
The mechanism realizes a three-axis rotational motion of the moving platform. 
Given a desired platform orientation $\mathbf{R}$, we compute the global positions of the two platform attachment points $\mathbf{C}$ and $\mathbf{D}$, and then solve the two crank angles $(\alpha,\beta)$ such that both rod-length constraints are satisfied.

\textbf{Pose parameterization and yaw decoupling.} Let $\{O\}$ denote the base frame and $\{M\}$ the moving-platform frame whose origin is located at the universal-joint center $M$.
The neck mechanism realizes a pure rotational motion; hence the position of $M$ is constant in $\{O\}$ and only the platform orientation varies. We parameterize the desired platform orientation by yaw--pitch--roll angles $(\psi,\theta,\varphi)$ and define the rotation matrix
\begin{equation}
\mathbf R(\theta,\varphi,\psi) = \mathbf R_x(\varphi)\,\mathbf R_y(\theta)\,\mathbf R_z(\psi),
\label{eq:app_R_def}
\end{equation}
where $\mathbf R_x(\cdot)$, $\mathbf R_y(\cdot)$, and $\mathbf R_z(\cdot)$ are the standard elementary rotations about the $x$, $y$, and $z$ axes, respectively.
In our design, yaw $\psi$ is actuated by an independent base motor and does not couple into the pitch/roll crank solving.
Equivalently, the two crank--rod chains can be solved in a yaw-compensated frame, and the IK mapping $(\theta,\varphi)\mapsto(\alpha,\beta)$ is independent of $\psi$.
Therefore, the following derivation uses the effective pitch--roll rotation $\mathbf R_{pr}(\theta,\varphi)$ (with yaw removed); for simplicity we keep the notation $\mathbf R$.

\textbf{Computing the attachment points $\mathbf C$ and $\mathbf D$.} Let ${}^{M}\mathbf c$ and ${}^{M}\mathbf d$ be the constant coordinates of the two upper attachment points $C$ and $D$ expressed in $\{M\}$.
Their global positions are
\begin{equation}
\mathbf C = \mathbf M + \mathbf R\,{}^{M}\mathbf c,\qquad
\mathbf D = \mathbf M + \mathbf R\,{}^{M}\mathbf d,
\label{eq:app_CD_global}
\end{equation}
where $\mathbf M$ is the (constant) position of the platform origin in the base frame.

\textbf{Left chain (solving $\alpha$).}
Let $\mathbf O_1$ be the left motor center.
Point $A$ is the lower spherical joint driven by the left crank, with crank radius $r$ and crank angle $\alpha$.
Assuming the crank rotates in the local $y$--$z$ plane about the $x$ axis, the position of $A$ is
\begin{equation}
\mathbf A(\alpha) = \mathbf O_1 +
\begin{bmatrix}
0\\ r\cos\alpha\\ r\sin\alpha
\end{bmatrix}.
\label{eq:app_A_of_alpha}
\end{equation}
The left rod length is constant, $\|\mathbf C-\mathbf A\| = l_1$, yielding
\begin{equation}
\|\mathbf C-\mathbf O_1-\mathbf r_1(\alpha)\|^2 = l_1^2,
\qquad
\mathbf r_1(\alpha):=
\begin{bmatrix}
0\\ r\cos\alpha\\ r\sin\alpha
\end{bmatrix}.
\label{eq:app_left_length_constraint}
\end{equation}
Define $\Delta \mathbf C := \mathbf C-\mathbf O_1 = [\Delta x,\Delta y,\Delta z]^\top$.
Expanding \eqref{eq:app_left_length_constraint} and using $\|\mathbf r_1\|^2=r^2$ gives the scalar constraint
\begin{equation}
\Delta y\cos\alpha + \Delta z\sin\alpha = Z_1, 
\label{eq:app_left_scalar}
\end{equation}
\begin{equation}
Z_1 := \frac{\Delta x^2+\Delta y^2+\Delta z^2 + r^2 - l_1^2}{2r}.
\end{equation}
Introduce the auxiliary quantities
\begin{equation}
K_1 := \sqrt{\Delta y^2+\Delta z^2},\qquad
\phi_1 := \operatorname{atan2}(\Delta y,\Delta z),
\label{eq:app_left_aux}
\end{equation}
so that $\Delta y = K_1\sin\phi_1$ and $\Delta z = K_1\cos\phi_1$.
Then \eqref{eq:app_left_scalar} becomes
\begin{equation}
K_1\sin(\alpha+\phi_1)=Z_1.
\label{eq:app_left_aux_eq}
\end{equation}
A real solution exists if
\begin{equation}
\left|\frac{Z_1}{K_1}\right|\le 1
\quad\text{and}\quad K_1>0.
\label{eq:app_left_feasible}
\end{equation}
The two closed-form solution branches are
\begin{equation}
\alpha_1 = \arcsin\!\left(\frac{Z_1}{K_1}\right) - \phi_1,
\qquad
\alpha_2 = \pi - \arcsin\!\left(\frac{Z_1}{K_1}\right) - \phi_1.
\label{eq:app_alpha_solutions}
\end{equation}
If $K_1=0$, then $\Delta y=\Delta z=0$ and \eqref{eq:app_left_scalar} reduces to $0=Z_1$, which is either infeasible ($Z_1\neq 0$) or underdetermined ($Z_1=0$); this singular case does not occur in the nominal workspace.

\textbf{Right chain (solving $\beta$).}
The right chain admits an identical derivation to the left chain and is therefore only summarized here.
Let $\mathbf{O}_2$ be the right motor center and $\mathbf{D}$ the corresponding platform attachment point.
Define
\begin{equation}
\Delta \mathbf{D} := \mathbf{D}-\mathbf{O}_2 = [\bar\Delta x,\bar\Delta y,\bar\Delta z]^\top,
\label{eq:app_deltaD_def}
\end{equation}
and denote the crank radius and rod length by $r$ and $l_2$, respectively.
With the crank rotating in the local $y$--$z$ plane, the rod-length constraint reduces to the scalar equation
\begin{equation}
\bar\Delta y\cos\beta + \bar\Delta z\sin\beta = Z_2, 
\label{eq:app_right_scalar}
\end{equation}
\begin{equation}
    Z_2 := \frac{\bar\Delta x^2+\bar\Delta y^2+\bar\Delta z^2 + r^2 - l_2^2}{2r}.
\end{equation}
Introduce the auxiliary quantities
\begin{equation}
K_2 := \sqrt{\bar\Delta y^2+\bar\Delta z^2},\qquad
\phi_2 := \operatorname{atan2}(\bar\Delta y,\bar\Delta z),
\label{eq:app_right_aux_compact}
\end{equation}
which yields $K_2\sin(\beta+\phi_2)=Z_2$.
A real solution exists if $\left|Z_2/K_2\right|\le 1$ (with $K_2>0$), and the two closed-form branches are
\begin{equation}
\beta_1 = \arcsin\!\left(\frac{Z_2}{K_2}\right) - \phi_2,\qquad
\beta_2 = \pi - \arcsin\!\left(\frac{Z_2}{K_2}\right) - \phi_2.
\label{eq:app_beta_solutions}
\end{equation}

Equations \eqref{eq:app_alpha_solutions} and \eqref{eq:app_beta_solutions} each admit two branches corresponding to the two assembly modes of a planar crank--rod linkage.
In control, we select the branch that (i) satisfies the mechanical joint limits and (ii) is closest to the previous command to ensure smooth motion.

\subsubsection{Architecture matters}
\label{sec:app-architecture-matters}
To examine how mechanism architecture affects scalability, we additionally apply the same automatic design algorithm to an open-source robotic head, Morpheus~\cite{zhang2025morpheus}. Morpheus decomposes facial actuation into multiple modules. The eyebrow module separates the eyebrow-head and eyebrow-peak motions, implemented by two four-bar linkages. The eye module couples the two eyeballs into a shared 2-DoF gaze mechanism and includes eyelid DoFs to approximate realistic blinking. The mouth module discretizes the lips into eight control points, where the mouth corners are driven by planar five-bar mechanisms and the remaining lip points control opening/closing; the jaw further supports lateral translation via a rack-and-pinion mechanism.

We then run our optimization on Morpheus under the same protocol. Since Morpheus contains additional tendon-driven DoFs that are not compatible with our linkage-based parametric synthesis, we conservatively exclude the tendon components and optimize only the linkage substructures. Under this setting, the success rate is 13.3\% (2/15), indicating that the remaining mechanical couplings and geometric complexity are not well aligned with scalable, constraint-driven automatic synthesis. In contrast, our head architecture is explicitly designed around modular, semantically partitioned linkage actuation, which achieves comparable expressiveness with a simpler and more scalable structure, making it amenable to robust automatic design.

\subsection{Optimization Algorithm Details}
\label{sec: app-optimization details}

\subsubsection{Initialization of geometric parameters and layout}
\label{sec: app-coarse alignment}
To ensure a physically plausible starting point for the hierarchical optimization, we analytically derive the initial spatial pose and geometric parameters of each module directly from the reconstructed 3D semantic landmarks.

\textbf{Eyebrow module.} We compute a vertical scaling ratio $\beta = H_\mathrm{mesh} / H_\mathrm{tpl}$, where $H_\mathrm{mesh}$ and $H_\mathrm{tpl}$ denote the characteristic facial height of the reconstructed mesh and the hardware template, respectively. Leveraging the anthropometric observation that relative vertical proportions are largely preserved, the initial spatial position of the eyebrow module base is determined by scaling the template's nominal eye-to-brow offset: $\mathbf{p}_\mathrm{brow}^\mathrm{init} = \mathbf{p}_\mathrm{eye} + \beta \cdot (\mathbf{p}_\mathrm{brow}^\mathrm{tpl} - \mathbf{p}_\mathrm{eye}^\mathrm{tpl}).$ We only initialize the module pose in the global head frame while keeping the linkage lengths unchanged for subsequent optimization. 

\textbf{Eyes module.} We use the semantic eye landmarks, each providing six 3D points $\{\mathbf{p}_{i} \in \mathbb{R}^{3}\}$ on the eye contour. Modeling the eye as a spherical joint, we employ a least-squares sphere fitting algorithm to estimate the optimal geometric center $\mathbf{p}_\mathrm{eye} \in \mathbb{R}^{3}$ and the fitting radius $r_\mathrm{fit}$ based on these six spatial points. The inter-ocular distance is $\|\mathbf{c}_{L} - \mathbf{c}_{R}\|_{2}$. For fabrication, we compensate for the outer silicone skin and eyelid shell by setting the physical eyeball radius as $r_\mathrm{eye} = r_\mathrm{fit} - (t_\mathrm{skin} + t_\mathrm{lid} + \delta)$, where $t_\mathrm{skin}$ and $t_\mathrm{lid}$ denote the thickness of the silicone skin and the rigid eyelid shell, and $\delta$ represents the necessary assembly clearance. This analytic process fully constrains the spatial position and geometric dimensions of the eye mechanism.

\textbf{Mouth module.}
The mouth is initialized as four coupled sub-mechanisms: two symmetric four-bar linkages for the upper and lower lips, and two five-bar linkages for the left and right mouth corners. Similar to the eyebrow module, this coarse initialization stage only adjusts the 3D placement of each sub-mechanism in the global head frame, while keeping all linkage lengths identical to the hardware template. Specifically, for the lip mechanisms, we align their skin attachment points (the output point $E$ of each four-bar) to the semantic landmarks at the upper- and lower-lip midpoints, which directly determines the initial base placement of the corresponding lip submodules. For the mouth-corner mechanisms, we align the terminal skin attachment point $K$ of each five-bar to the left/right mouth-corner landmarks, thereby fixing the initial placement of the two corner submodules. 

\textbf{Jaw module.}
The jaw mechanism is not optimized in our hierarchy and is fixed after coarse initialization. We assume that the relative spatial relationship between the jaw module and the eye module is consistent across different heads (the jaw base is positioned by a fixed template offset with respect to the eye center, up to a global facial scale). Accordingly, we reuse the same scale-aware placement rule as in the eyebrow initialization. The initial jaw base position is computed by scaling the template eye-to-jaw offset:
$\mathbf{p}_\mathrm{jaw}^\mathrm{init}
=
\mathbf{p}_\mathrm{eye}
+
\beta \cdot
\big(\mathbf{p}_\mathrm{jaw}^\mathrm{tpl}
-
\mathbf{p}_\mathrm{eye}^\mathrm{tpl}\big).$
Since the jaw module only needs to produce a one-DoF opening/closing motion, this coarse placement is sufficient for our purposes; we therefore keep the jaw mechanism fixed during subsequent optimization and exclude it from the inner-loop kinematic synthesis.

\begin{figure}
    \centering
    \includegraphics[width=1.0\linewidth]{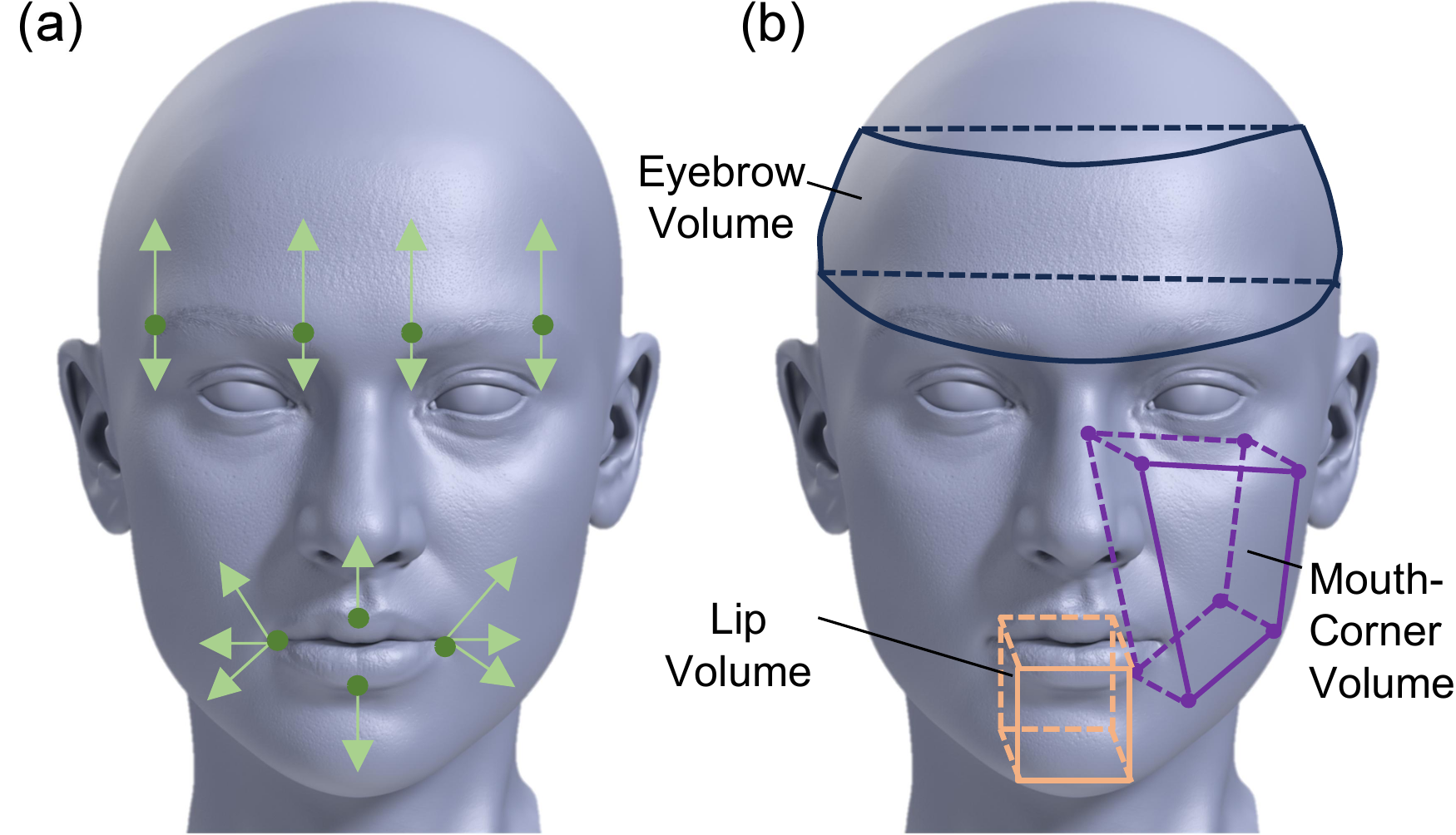}
    \caption{(a) \textbf{AU-guided target motions.} Canonical AU-derived motion directions for the optimized facial attachment points (eyebrow, lips, and mouth corners), defining the target trajectory primitives used in the inner-loop synthesis. (b) \textbf{Anatomy-guided feasible volumes.} Region-specific anatomically plausible feasible volumes that bound the optimizer’s search space, including the eyebrow volume, lip volume, and mouth-corner volume.}
    \label{fig:app-2D traj/3D volume}
    \vspace{-0.5cm}
\end{figure}

\subsubsection{Determine anatomy-guided feasible motion volumes}
\label{sec: app-Determine local 3D design space.}
We instantiate the anatomy-guided feasible motion volumes only for the actively optimized regions (eyebrow and mouth), providing explicit geometric bounds used to construct $\Omega_\mathbf{feas}$ (Fig.~\ref{fig:app-2D traj/3D volume}(b)).

\textbf{Eyebrow volume.} Following the anthropometric evidence in~\cite{chong2021three}, we bound the eyebrow workspace by two parallel planes along the local vertical direction. Concretely, taking the lowest eyebrow landmark as a reference, we set a lower plane 10 $mm$ below this point and an upper plane 50 $mm$ above it. The feasible eyebrow motion volume is defined as the region enclosed between these two planes (together with the local facial surface boundary), ensuring that the synthesized brow trajectories remain within a bio-plausible elevation range.

\textbf{Mouth-corner volume.} For the mouth corner, we follow the facial anatomy described in \cite{drake2015grays} and construct a cheek plane using three salient landmarks: the mouth-corner point, the zygomatic (cheekbone) high point, and the mandibular angle. We then extrude this plane inward (toward the skull interior) by 50 $mm$ to form a bounded slab, which serves as the feasible volume for the mouth-corner linkage end-effector. This construction prevents unrealistic outward penetration while providing sufficient depth for embedding the corner mechanism within the facial shell.

\textbf{Lip volume.} For the upper and lower lips, we define a thin mid-sagittal corridor to reflect the anatomically constrained motion near the mouth center. Using the head’s vertical symmetry plane as the reference, we restrict the lip attachment points to remain within ±10 $mm$
 laterally from this plane. In addition, we impose an upper bound of 30 $mm$ above the neutral lip position, limiting excessive upward displacement while still allowing the optimizer to scale vertical opening/closing trajectories to the largest feasible amplitude.

 Together, these region-specific bounds provide conservative but bio-plausible envelopes for the inner-loop synthesis, regularizing the search while retaining sufficient freedom for maximizing expressiveness under collision-free assembly constraints.

\subsubsection{Determine AU-derived trajectories}
\label{sec: app-Determine trajectory}
In the inner-loop optimization, we impose trajectory constraints only on the actively optimized facial modules (eyebrow, lips, and mouth corners), since the eye and jaw modules are fixed after coarse initialization. Specifically, we derive a set of canonical motion trajectories from AU-based facial motion analysis and use them as kinematic targets for synthesis (Fig.~\ref{fig:app-2D traj/3D volume}(a)). 

For the eyebrow module, the two skin attachment points corresponding to the inner brow and brow peak are constrained to follow \emph{vertical} trajectories, reflecting the dominant up/down motion patterns of these AUs. Similarly, for the upper- and lower-lip modules, the attachment points located at the lip midpoints are also constrained to move along the \emph{vertical} direction, which captures the primary opening/closing deformation at the mouth center. For the mouth-corner module (five-bar linkage), we constrain the planar end-effector motion with three representative trajectories that realize upward raising, lateral translation, and downward pulling of the mouth corner, respectively.

During subsequent planning and optimization, each canonical trajectory is automatically assigned a scalar amplitude (scaling) factor $\alpha_k$. The optimizer selects these scaling factors to maximize expressiveness while satisfying kinematic feasibility and interference constraints, ensuring that the synthesized mechanisms achieve the largest realizable motion range consistent with the AU-derived motion directions.

\subsubsection{Optimization parameters}
\label{sec: app-Optiomization parameters}
We summarize the key hyperparameters used in our hierarchical automatic design optimization in Tab.~\ref{tab:design_hyperparams}, including settings for the inner loop, the outer-loop collision-driven refinement, and the L-BFGS solver.

\begin{table}[t]
\centering
\footnotesize
\setlength{\tabcolsep}{4pt}
\renewcommand{\arraystretch}{1.05}
\caption{Hyperparameters for hierarchical optimization.}
\resizebox{\linewidth}{!}{
\begin{tabular}{l p{0.63\linewidth} c}
\toprule
Hyperparameter & Meaning & Value \\
\midrule
$\beta$ & Global mesh scaling factor used to bring the reconstructed head to metric scale. & per-head \\
\midrule
$K$ & Number of canonical AU-derived target trajectories. & 8 \\
$M$ & Number of sampled waypoints per trajectory (also used for collision checking along motion). & 10 \\
$\alpha_{\min}$ & Lower bound on amplitude coefficient $\alpha_k$ to avoid numerical instability. & 0.1 \\
$\alpha^{\mathrm{limit}}_{\mathrm{init}}$ & Initial amplitude limit $\alpha_k^{\mathrm{limit}}$ before outer-loop scheduling. & 2.0 \\
\midrule
$\omega_{\mathrm{amp}}$ & Weight for amplitude term $\mathcal{L}_{\mathrm{amp}}$ (encourages larger expression range). & 1.0 \\
$\omega_{\mathrm{fit}}$ & Weight for trajectory fitting term $\mathcal{L}_{\mathrm{fit}}$ (tracking accuracy). & 10 \\
$\omega_{\mathrm{man}}$ & Weight for manipulability term $\mathcal{L}_{\mathrm{man}}$ (avoids kinematic singularities / dead zones). & 0.1 \\
\midrule
$S$ & Maximum number of outer-loop refinement iterations. & 10 \\
$\epsilon$ & Collision-resolution clearance in QP constraints. & 0.5\,mm \\
$D_{\max}$ & Maximum accepted base-pose update magnitude $\|\Delta\mathbf{p}_k\|_2$ before triggering amplitude scheduling. & 3.0\,mm \\
$\gamma$ & Amplitude decay factor for scheduling when $\|\Delta\mathbf{p}_k\|_2>D_{\max}$. & 0.9 \\
\midrule
L-BFGS iters & Maximum iterations for each inner-loop solve. & 100 \\
L-BFGS tol & Termination tolerance (gradient norm / relative improvement). & $10^{-6}$ \\
\bottomrule
\end{tabular}
}
\label{tab:design_hyperparams}
\end{table}

\subsubsection{Baselines}
\label{sec: app-Optiomization baselines}
To rigorously evaluate the efficacy of our proposed hierarchical optimization framework, we implemented three comparative baselines. This part details the mathematical formulation and algorithmic procedures for each baseline.

\textbf{Local-only optimization (Ours w/o outer loop).}
This baseline evaluates the necessity of the outer assembly refinement loop. It performs optimization strictly within independent modules, ignoring global inter-module constraints. This approach implicitly assumes that $V_\mathrm{int}(\Omega_i, \Omega_j) = 0$. However, in practice, components optimized locally frequently encroach upon the space required by adjacent modules.

\textbf{Joint optimization (Global-Opt).} 
This baseline represents a monolithic optimization strategy that attempts to solve for both the internal mechanism parameters and the global assembly layout simultaneously. Unlike our hierarchical framework which decouples mechanism synthesis (inner loop) from assembly refinement (outer loop), this baseline merges all objectives into a single unconstrained loss function.

We combine the original mechanism synthesis objectives ($\mathcal{L}_\mathrm{mech}$) with the global assembly constraints (collision and position retention). Let $\mathbf{X} = \{ \mathbf{\Phi}, \mathbf{\Theta}, \boldsymbol{\alpha}, \mathbf{p}_\mathrm{base} \}$ denote the full set of optimization variables, including mechanism parameters and the base positions of all modules. The joint objective function is defined as:
\begin{equation}
    \min_{\mathbf{X}} \quad \mathcal{L}_\mathrm{joint} = \mathcal{L}_\mathrm{mech} + \omega_\mathrm{pos} \mathcal{L}_\mathrm{pos} + \omega_{coll} \mathcal{L}_\mathrm{coll}.
\end{equation}
\begin{itemize}
    \item Mechanism Synthesis Term ($\mathcal{L}_\mathrm{mech}$):
This term preserves the original kinematic objectives defined in Eq. (\ref{eq:innerloop_overall_obj_constrained}), aiming to optimize amplification, fitting accuracy, and manipulability:
\begin{equation}
    \mathcal{L}_\mathrm{mech} = \sum_{k=1}^K \left( \omega_\mathrm{amp}\mathcal{L}_\mathrm{amp}^{(k)} + \omega_\mathrm{fit}\mathcal{L}_\mathrm{fit}^{(k)} - \omega_\mathrm{man}\mathcal{L}_\mathrm{man}^{(k)} \right),
\end{equation}
where $k$ indexes the mechanism modules.

    \item Assembly Constraint Terms:
Crucially, the hard geometric constraints $\mathbf{P}(\boldsymbol{\theta}_{k,m}) \subset \Omega_{\mathrm{feas}}$ (feasibility within the shell) and inter-module non-intersection constraints are relaxed into soft penalty terms: (i) Position Retention ($\mathcal{L}_{pos}$): Ensures the mechanism bases remain close to their initialized template positions $\mathbf{p}_\mathrm{base}^\mathrm{init}$:
\begin{equation}
     \mathcal{L}_\mathrm{pos} = \sum_{k=1}^{K} \| \mathbf{p}_{\mathrm{base}, k} - \mathbf{p}_{\mathrm{base,} k}^\mathrm{init} \|^2_2.
\end{equation}
(ii) Global Collision Penalty ($\mathcal{L}_\mathrm{coll}$): we leverage a collision detection API to identify the set of colliding pairs $\mathcal{C}_\mathrm{active}$ at each iteration. Let $V_\mathrm{int}(\Omega_i, \Omega_j)$ be the intersection volume directly returned by the physics engine query for components $i$ and $j$. The collision loss is simply the sum of these squared interference magnitudes:
\begin{equation}
      \mathcal{L}_{coll} = \sum_{(i, j) \in \mathcal{C}_\mathrm{active}} \left( V_\mathrm{int}(\Omega_i, \Omega_j) \right)^2.
\end{equation}
Here, $\mathcal{C}_\mathrm{active}$ denotes the set of all component pairs currently in collision. If no collision is detected, $\mathcal{C}_\mathrm{active} = \emptyset$ and $\mathcal{L}_\mathrm{coll} = 0$.
\end{itemize}

\begin{figure*}
    \centering
    \includegraphics[width=1.0\linewidth]{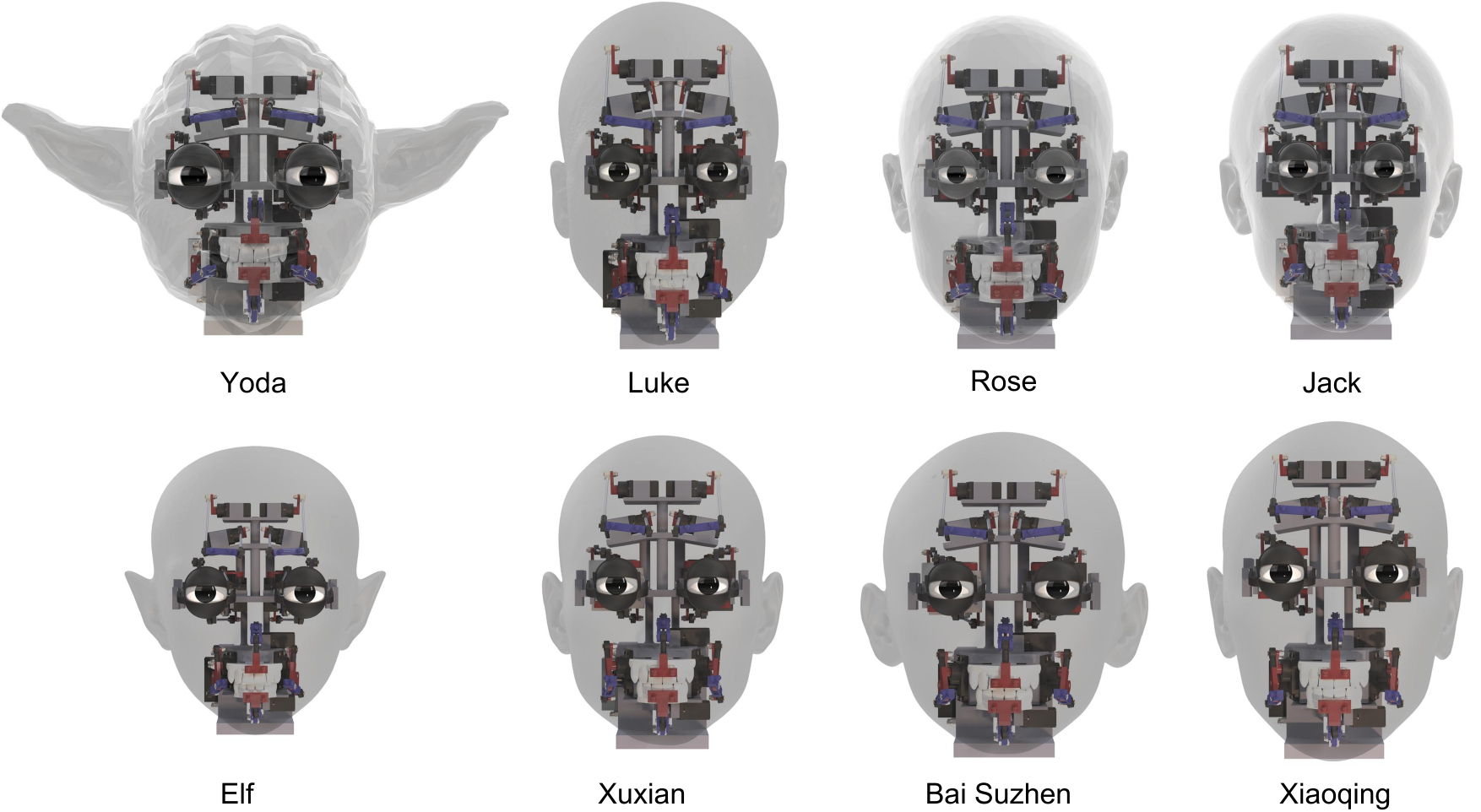}
    \caption{\textbf{Automated mechanism synthesis across a spectrum of facial morphologies.} We visualize the optimized internal CAD assemblies for eight distinct identities, rendered at relative physical scale to highlight volumetric diversity. It spans cinematic characters (\textit{Yoda}, \textit{Luke}, \textit{Rose}, \textit{Jack}), a stylized \textit{Elf}, and characters from Chinese folk (\textit{Xu Xian}, \textit{Bai Suzhen}, \textit{Xiaoqing}).}

    \label{fig:diverse synthesis}
\end{figure*}

\textbf{Heuristic adjustment (Local + Heuristic).}
This baseline incorporates an outer assembly loop to resolve global collisions but employs a heuristic repulsion strategy instead of our rigorous MTV-based  projection.

First, we identify the set of active collision constraints. For a component $k$, let $\mathcal{C}_k$ be the set of obstacles it currently intersects. Second, for each collision $j \in \mathcal{C}_k$, we calculate a repulsion vector $\mathbf{v}_\mathrm{rep}$ based on centroids $\mathbf{c}_k, \mathbf{c}_l$:
\begin{equation}
    \mathbf{v}_\mathrm{rep}^{(j)} = \frac{\mathbf{c}_k - \mathbf{c}_l}{\| \mathbf{c}_k - \mathbf{c}_l \| + \epsilon}.
\end{equation}
We resolve the collisions by applying a fixed-step displacement. If a component collides with multiple obstacles, we apply the superposition of all repulsion vectors:
\begin{equation}
    \mathbf{p}_k^{(s+1)} \leftarrow \mathbf{p}_k^{(s)} + \gamma \sum_{j \in \mathcal{C}_k} \mathbf{v}_\mathrm{rep}^{(j)}.
\end{equation}

\subsubsection{Diverse facial mechanism synthesis visualization}
To strictly validate the geometric versatility of our hierarchical optimization framework, we conducted a batch synthesis experiment on a diverse dataset of eight 3D portraits (Fig.~\ref{fig:diverse synthesis}). This set encompasses both photorealistic human subjects (Luke, Rose, Jack, Xu Xian, Bai Suzhen, and Xiaoqing from \textit{The Legend of the White Snake}), and stylized non-human characters with extreme morphological constraints (Yoda, Elf). Crucially, the visualization confirms that our algorithm successfully generates collision-free, valid linkage assemblies for every target face. The results confirm that the system maintains kinematic integrity while accommodating geometric variations, ranging from the flattened cranial structure of Yoda to the anthropometrically slender proportions of Xiaoqing.

\subsubsection{Failure cases}
Although our hierarchical pipeline is robust across a wide range of face geometries, it can fail when the target morphology violates the feasibility assumptions of the fixed mechanism template.

\textbf{Template-infeasible morphology.}
In some extreme geometries, the design becomes \emph{structurally infeasible} under our current mechanism template. For instance, heads with a very pointed chin and substantially reduced internal clearance relative to the template leave insufficient volume to place the mouth-corner and lower-lip linkages while satisfying kinematic limits and collision constraints. Since our pipeline keeps the mechanism topology fixed and only optimizes continuous parameters (base poses and link lengths), feasibility cannot be recovered when the constraint set is empty for the given template. In such cases, optimization terminates at the iteration budget with persistent interferences and/or severely contracted amplitude limits, suggesting that topology changes (rather than parameter tuning) would be required to obtain a valid design.

\textbf{Non-resolvable multi-module interlocking.}
A second failure mode arises from coupled collisions involving multiple modules, where interferences are not removable by small base-pose adjustments alone. In these cases, linkages become interlocked with adjacent structures (e.g., skull/skin or neighboring mechanisms) across a range of configurations, and resolving one contact often introduces another. Consequently, the set of MTV constraints can become mutually inconsistent, yielding contradictory separation normals and an empty or near-empty feasible region for the QP. Since our outer loop only performs pose-level corrections and amplitude scheduling, it cannot handle collision patterns that fundamentally require re-parameterizing the mechanism structure or changing the linkage topology. As a result, optimization reaches the iteration budget with non-zero intersection volume.

\subsection{Manufacture Details}
\label{sec: app-manufacture details}
\subsubsection{3D printed materials}
All rigid mechanical components and the supporting shell are fabricated using Polylactic Acid (PLA) on a commercial high-speed FDM printer (Bambu Lab P1S). To ensure precise assembly and low-friction joint articulation without extensive post-processing, we explicitly incorporate a design clearance of $0.1$--$0.2$,mm into the CAD models for all moving interfaces. The manufacturing time reported in Tab. \textcolor{blue}{IV} 
is calculated as the cumulative sum of the estimated print durations for all component batches, assuming sequential production on a single printer unit.

\subsubsection{Silicone face skin}
The facial skin is fabricated using a double-mold casting technique. We utilize a standard RTV-2 molding silicone. The fabrication process proceeds as follows:
\textbf{Mold Preparation:} A rigid outer mold (defining the facial surface) and an inner core (defining the interface with the mechanical skull) are 3D printed and treated with a release agent.
\textbf{Degassing:} The two-component liquid silicone (Parts A and B) is mixed in a 1:1 ratio. To ensure a flawless surface texture free of air bubbles, the mixture undergoes vacuum degassing in a chamber at -0.1 MPa for approximately 10 minutes until all trapped air is evacuated.
\textbf{Casting and Curing:} The degassed mixture is slowly poured into the interstitial space between the assembled inner and outer molds. The skin is allowed to cure at room temperature (25$^\circ$C) for 4 hours before demolding, ensuring stable mechanical properties and precise geometric fidelity.

\subsubsection{Actuator choice and noise testing}
To balance torque density with the spatial constraints of the compact cephalic volume, we select GuoHua 9g digital micro-servos (180$^\circ$ range). The neck mechanism, which requires a higher holding torque, is driven by three NEMA 17 stepper motors.
Precise motion control is achieved via cascaded PCA9685 Pulse Width Modulation (PWM) drivers communicating with the central Orin computer over the $\mathrm{I}^2\mathrm{C}$ protocol. Crucially, to ensure temporal coherence between the digital generation pipeline and physical execution, the control loop operates at a fixed frequency of 30 Hz. This update rate is strictly synchronized with the frame rate of the upstream digital avatar system.

Acoustic emission is a critical metric for proximal Human-Robot Interaction, as excessive mechanical noise can disrupt user immersion and speech intelligibility. We analyze the spectral characteristics of the servo noise, identifying that the primary acoustic source is the high-velocity impact of the first-stage gear reduction. To mitigate this, we utilize high-precision gear manufacturing with optimized meshing tolerances, significantly reducing high-frequency gear whine compared to standard hobbyist servos.

We quantify the system's acoustic footprint using a calibrated decibel meter placed at a typical interaction distance of 0.5 m. The idle noise floor is measured at \textbf{30 dB}. According to acoustic standards for communication (ANSI S12.60), typical conversational speech levels range between 60--65 dB. Consequently, during active dialogue, the robot's generated speech output induces a robust auditory masking effect, rendering the mechanical actuation noise perceptually negligible and ensuring an undistracted user experience.

\subsection{Talking Head Synthesis Details}
\label{sec: app-talking head synthesis}
\subsubsection{Network architecture}
In this section, we elucidate the comprehensive architectural specifications of our Unified Dual-Speaker Interaction Framework (Fig.~\ref{fig:initeraction synthesis and mapping}(a)).

\textbf{Dual-speaker joint encoder.} The conversational motion synthesis framework is underpinned by a Dual-Speaker Joint Encoder designed to capture reciprocal acoustic dynamics. The audio processing branch operates on raw waveforms sampled at 16 $\mathrm{kHz}$. We leverage the robust feature extraction capabilities of the pre-trained Wav2Vec 2.0~\cite{baevski2020wav2vec} model as the acoustic backbone. Structurally, this encoder comprises a stack of 12 Transformer layers with a hidden embedding dimension of 1024. To bridge the domain gap between acoustic and visual modalities, the extracted features are subsequently passed through a learnable linear projection layer.
Following the feature extraction and linear projection (Eq. \textcolor{blue}{3}), 
we obtain two parallel latent audio streams $\mathbf{Z}_A$ and $\mathbf{Z}_B$ representing the target robot (Speaker A) and the interlocutor (Speaker B), respectively. 
Before integrating the two streams, we add learnable temporal positional encodings ($\mathbf{P}$) and identity-specific style embeddings ($\mathbf{S}$) to the raw features to obtain the enriched representations $\tilde{\mathbf{Z}}_A$ and $\tilde{\mathbf{Z}}_B$:
\begin{equation}
\tilde{\mathbf{Z}}_A = \mathbf{Z}_A + \mathbf{P} + \mathbf{S}_A, \qquad
\tilde{\mathbf{Z}}_B = \mathbf{Z}_B + \mathbf{P} + \mathbf{S}_B.
\end{equation}

\textbf{Turn-aware audio gating.} To construct a unified context vector that reflects the immediate conversational state, we employ a \textbf{Turn-Aware Gating} mechanism.
Let $g_A(t) \in [0, 1]$ denote the speech activity probability of Speaker A at frame $t$, derived from the audio intensity. We synthesize a turn-conditioned memory $\mathbf{M}_A \in \mathbb{R}^{T \times d}$ by dynamically interpolating between the two streams:
\begin{equation} 
\mathbf{M}_A(t) = g_A(t) \cdot \tilde{\mathbf{Z}}_A(t) + \big(1 - g_A(t)\big) \cdot \tilde{\mathbf{Z}}_B(t). 
\label{eq: turn_gate} 
\end{equation}
This gating mechanism forces the network to attend to the target's own speech for lip-sync generation when active ($g_A \approx 1$) and shift attention to the interlocutor's audio for reactive listening behaviors when passive ($g_A \approx 0$).

\textbf{Transformer backbone.} The core sequence modeling is performed by a Transformer architecture comprising an Encoder and a Decoder, both configured with $N=3$ layers.
\begin{itemize}
\item \textbf{Encoder:} The encoder processes the gated memory $\mathbf{M}_A$ to capture long-range temporal dependencies. Each layer employs a 4-head self-attention mechanism with a hidden dimension of $d=256$ and a Feed-Forward Network (FFN) with an expansion dimension of 512.
\item \textbf{Turn-Aware Alignment Attention:} To mitigate motion jitter during the critical transitions between speaking and listening roles, we introduce a specialized attention layer between the encoder and decoder. Unlike standard cross-attention, this layer modulates the attention weights using a temporal bias matrix $\mathbf{B}_\mathbf{turn}$ derived from the gradient of the turn signal $\nabla g_A(t) = |g_A(t) - g_A(t-1)|$:
\begin{equation}
    \text{Attention}(\mathbf{Q}, \mathbf{K}, \mathbf{V}) = \text{softmax}\left( \frac{\mathbf{Q}\mathbf{K}^\top}{\sqrt{d}} - \lambda \cdot \mathbf{B}_\mathbf{turn} \right) \mathbf{V}.
\end{equation}
This mechanism explicitly suppresses attention noise at turn boundaries, ensuring smooth kinematic transitions as the robot switches from lip-syncing to reactive nodding.

\item \textbf{Decoder:} The decoder follows a similar 3-layer structure ($d=256$, 4 heads) but incorporates additional cross-attention layers to integrate the encoded contextual features with $T$ learnable motion queries equipped with sinusoidal positional encodings.

\end{itemize}

\textbf{Expression Modulation and Output.} The decoded features are passed through an MLP to regress the final kinematic parameters. This module consists of a two-layer perceptron: the first layer expands the feature space to 512 dimensions, followed by Layer Normalization and a ReLU activation; the second layer projects back to 256 dimensions. Finally, a linear regression head maps the modulated features to the output vector $\hat{\mathbf{Y}}_t \in \mathbb{R}^{55}$.
The output comprises 52 ARKit blendshape coefficients for facial deformation and $3$ Euler angles representing the head pose (pitch, yaw, roll). We apply a Sigmoid activation function $\sigma(\cdot)$ exclusively to the blendshape branch to strictly bound the coefficients within the valid physical range of $(0, 1)$, ensuring plausible facial deformations.

\subsubsection{Loss function}
We train the network to predict a temporally coherent sequence of motion parameters from dual-speaker audio. Each frame is parameterized by a 55-D vector $\mathbf{y}_t=[\mathbf{b}_t;\mathbf{r}_t]\in\mathbb{R}^{55}$, where $\mathbf{b}_t\in\mathbb{R}^{52}$ are blendshape coefficients, and $\mathbf{r}_t\in\mathbb{R}^{3}$ denotes head rotation.
Our objective is two-fold: (i) accurately reconstruct the ground-truth blendshape parameters at each frame, and (ii) enforce smooth and natural temporal dynamics by suppressing frame-to-frame jitter. 
To this end, we optimize a weighted sum of a reconstruction loss and a velocity-consistency loss.

Let $\hat{\mathbf{Y}}=[\hat{\mathbf{y}}_1,\dots,\hat{\mathbf{y}}_T]^\top \in \mathbb{R}^{T\times 55}$ denote the predicted blendshape parameters and $\mathbf{Y}=[\mathbf{y}_1,\dots,\mathbf{y}_T]^\top \in \mathbb{R}^{T\times 55}$ the ground truth, where $T$ is the number of animation frames.
We supervise per-frame accuracy using an $\ell_2$ loss:
\begin{equation}
\mathcal{L}_{\mathrm{rec}}
=
\frac{1}{T}\sum_{t=1}^{T}
\left\|
\hat{\mathbf{y}}_t-\mathbf{y}_t
\right\|_2^2.
\label{eq:loss_rec_app}
\end{equation}

To encourage smooth facial motions and reduce high-frequency artifacts, we additionally match the first-order temporal differences (velocities) between the prediction and the ground truth:
\begin{equation}
\hat{\mathbf{v}}_t = \hat{\mathbf{y}}_{t} - \hat{\mathbf{y}}_{t-1},\qquad
\mathbf{v}_t = \mathbf{y}_{t} - \mathbf{y}_{t-1},\qquad t=2,\dots,T,
\label{eq:vel_def_app}
\end{equation}
and define the velocity loss as
\begin{equation}
\mathcal{L}_{\mathrm{vel}}
=
\frac{1}{T-1}\sum_{t=2}^{T}
\left\|
\hat{\mathbf{v}}_t-\mathbf{v}_t
\right\|_2^2.
\label{eq:loss_vel_app}
\end{equation}

The final training objective is
\begin{equation}
\mathcal{L}
=
\lambda_{\mathrm{rec}}\mathcal{L}_{\mathrm{rec}}
+
\lambda_{\mathrm{vel}}\mathcal{L}_{\mathrm{vel}},
\label{eq:loss_total}
\end{equation}
where $\lambda_{\mathrm{rec}}$ and $\lambda_{\mathrm{vel}}$ balance reconstruction accuracy and temporal consistency.

\subsubsection{Training details}
The network parameters are optimized using the Adam algorithm, initialized with a constant learning rate of $1 \times 10^{-4}$. The training curriculum spans a total of 200 epochs, processing data in mini-batches of size 32. We employ a distributed training strategy on a server node equipped with 4 NVIDIA L20 GPUs. Under this computational configuration, the complete training pipeline requires approximately 20 hours to reach convergence.

\subsubsection{Evaluation metrics}
We provide detailed definitions of the quantitative metrics employed to assess the performance of both speaker-centric synchronization and listener-centric reactive behaviors.

\textbf{Lip Vertex Error (LVE, $\downarrow$).} Measures lip-synchronization accuracy by calculating the geometric deviation of the mouth region. Specifically, we compute the maximum Euclidean distance between the generated and ground-truth lip vertices for each frame, averaged over the entire sequence. Lower values indicate more precise mouth articulation.

\textbf{Mean Squared Error (MSE, $\downarrow$).} Measures the frame-wise reconstruction fidelity of facial expressions. It is defined as the mean squared error between the predicted expression representations (blendshape coefficients) and the ground truth over the time dimension. Lower values imply better preservation of the original performance.

\textbf{Pose Dynamic Deviation (PDD, $\downarrow$).} Measures whether the temporal dynamics of the generated head pose match the ground-truth distribution. We compute the standard deviation of the first-order temporal derivatives (velocities) for both sequences. The metric reports the $\ell_1$ distance between these statistics:
\begin{equation}
\text{PDD}=\left\|\mathrm{std}_t(\dot{\mathbf{r}}^{\text{gen}}_t)-\mathrm{std}_t(\dot{\mathbf{r}}^{\text{gt}}_t)\right\|_1,
\end{equation}
where $\dot{\mathbf{r}}$ denotes the head pose velocity. Lower values indicate that the robot's motion intensity and stability align with real human behaviors.

\textbf{SI for Diversity (SID, $\uparrow$).} Measures the richness of the generated listener behaviors. We cluster the generated motion features into $K$ discrete modes and compute the Shannon entropy of the resulting cluster assignment distribution $\mathcal{P} = \{p_1, \dots, p_K\}$:
\begin{equation}
\text{SID}=-\sum_{k=1}^{K}p_k\log p_k.
\end{equation}
Higher values indicate broader coverage of motion modes, suggesting that the model avoids mode collapse and generates non-repetitive reactions.

\textbf{Upper-face Dynamic Deviation (FDD, $\downarrow$).} 
Measures the realism of upper-face motion dynamics (eyebrows and eyelids). Similar to PDD, it is calculated by comparing the standard deviation of the temporal derivatives for the upper-face blendshape parameters $\mathbf{b}_{\text{upper}}$:
\begin{equation}
\text{FDD}=\left\|\mathrm{std}_t(\dot{\mathbf{b}}^{\text{upper}}_{\text{gen},t})
-\mathrm{std}_t(\dot{\mathbf{b}}^{\text{upper}}_{\text{gt},t})\right\|_1.
\end{equation}
Lower values indicate that the dynamic expressiveness of the upper face closely resembles the ground truth.

\subsubsection{Dataset details}
We build a dual-speaker interaction dataset from publicly available YouTube interview-style videos and RealTalk~\cite{ji2024realtalk}, which provide diverse face-to-face conversational behaviors. We select videos where (i) both speakers are clearly visible for most of the clip, (ii) the audio quality is sufficient for separation, and (iii) the conversation exhibits natural turn-taking dynamics. The selected videos are in 1920$\times$1080 resolution, recorded at 30 FPS, with audio resampled to 16 kHz. The selected videos span a wide range of languages and speaking styles, which later enables bilingual interactions in our system rather than restricting it to a single language.
After processing, the dataset contains 50 hours of conversation data, spanning 1,052 unique identities across 5,858 clips. Each clip contains on average 2.5 conversational turns, where speakers naturally alternate between speaking and listening roles. We split the data into training (5,282 clips) and testing (576 clips).

We start from long-form conversation videos and segment them into temporally coherent clips using TransNetV2~\cite{soucek2024transnet}. We keep segments with continuous face visibility for both participants and discard segments with heavy occlusions, extreme pose, or frequent shot changes. For each retained clip, we maintain a consistent speaker ordering (Speaker A/B) by tracking face detections over time and enforcing temporal continuity, ensuring that the extracted facial parameters and separated audio correspond to the same identity throughout the clip. 
For each clip, we apply SAM Audio~\cite{shi2025sam} to separate the mixed audio into two synchronized speaker streams, producing $\mathbf{A}_A$ and $\mathbf{A}_B$ at 16 kHz. We additionally apply basic quality checks to filter clips where separation is unreliable.
On the visual side, we run MediaPipe~\cite{Lugaresi2019MediaPipeAF} to obtain per-frame blendshape coefficients $\mathbf{b}\in\mathbb{R}^{52}$ and head rotation $\mathbf{R}\in\mathbb{R}^{3}$ for each speaker. We associate per-frame parameters to Speaker A/B using the tracked face identities, and we temporally align the facial parameters with the separated audio streams to form synchronized tuples.
The final dataset consists of synchronized tuples
$\mathcal{D}=\{(\mathbf{A}_A,\mathbf{A}_B,\mathbf{B}_A,\mathbf{B}_B,\mathbf{R}_A,\mathbf{R}_B)\}$,
where $\mathbf{B}$ stacks per-frame blendshape coefficients over time.

\subsection{Mapping Network Details}
\label{sec: app-mapping network}

\subsubsection{Baseline details} We provide implementation details for the baselines.

\textbf{Landmark representation.}
We extract $D{=}468$ 2D facial landmarks using MediaPipe~\cite{Lugaresi2019MediaPipeAF}.
We denote the landmark matrix as
$\mathbf{L}_t=\{(x_{t,j},y_{t,j})\}_{j=1}^{D}\in\mathbb{R}^{D\times 2}$,
and its vectorized form as
$\mathbf{l}_t=\mathrm{vec}(\mathbf{L}_t)\in\mathbb{R}^{2D}$.
We use $\mathbf{l}^H_t$ for the avatar/human landmarks and $\mathbf{l}^R_t$ for the robot landmarks.

\begin{itemize}
    \item \textbf{Landmark-NN (nearest-neighbor retrieval).}
Given a query landmark vector $\mathbf{l}^H_t$ at inference time, we retrieve the closest sample from the robot calibration set
$\{(\mathbf{l}^R_i,\mathbf{u}_i)\}_{i=1}^{N}$ and output its associated motor command:
\begin{equation}
    i^\star = \arg\min_{i\in\{1,\dots,N\}} \left\|\mathbf{l}^H_t-\mathbf{l}^R_i\right\|_2^2,
\qquad
\hat{\mathbf{u}}_t=\mathbf{u}_{i^\star}.
\end{equation}

\item \textbf{Landmark~\cite{chen2021smileeva} normalization + global MLP.}
We apply per-dimension min--max normalization to map human landmarks into the robot landmark range:
\begin{equation}
   \tilde{\mathbf{l}}^H
=
\left(\mathbf{l}^H-\mathbf{h}_{\min}\right)
\frac{\mathbf{r}_{\max}-\mathbf{r}_{\min}}{\mathbf{h}_{\max}-\mathbf{h}_{\min}}
+\mathbf{r}_{\min}, 
\end{equation}
$\mathbf{h}_{\min},\mathbf{h}_{\max}\in\mathbb{R}^{2D}$ and $\mathbf{r}_{\min},\mathbf{r}_{\max}\in\mathbb{R}^{2D}$ are per-coordinate minima/maxima computed over the human/avatar and robot landmark datasets, respectively.
We then train a 2-layer global MLP $\pi(\cdot)$ to regress motor commands with an MSE loss on the robot dataset:
\begin{equation}
    \mathcal{L}_{\text{MLP}}=\frac{1}{N}\sum_{i=1}^{N}\left\|\pi(\mathbf{l}^R_i)-\mathbf{u}_i\right\|_2^2.
\end{equation}
\item \textbf{Landmark~\cite{hu2024human} normalization + global MLP.}
Let $\mathbf{h}_s,\mathbf{r}_s\in\mathbb{R}^{2D}$ be the rest landmarks for human and robot, and define $\Delta=\mathbf{l}^H-\mathbf{h}_s$.
We normalize by scaling and clamping:
\begin{equation}
    \tilde{\mathbf{l}}^H
=
\mathbf{r}_s
+
 K \times
\mathrm{clip}\!\left(\Delta,\ \mathbf{r}_{\min}-\mathbf{r}_s,\ \mathbf{r}_{\max}-\mathbf{r}_s\right),
\end{equation}
where $K$ is a scaling vector.
We then apply the same global MLP and MSE objective to predict motor commands,
$\hat{\mathbf{u}}_t=\pi(\tilde{\mathbf{l}}^H_t)$.

\end{itemize}

\textbf{FLAME~\cite{li2017learning} representation.} FLAME defines a deformable head mesh as a function of identity shape, facial expression, and articulated pose:
\begin{equation}
    \mathcal{M}(\theta, \beta, \psi) \rightarrow \mathbf{V} \in \mathbb{R}^{N \times 3}
\end{equation}
where $\beta$ are PCA-based coefficients that define identity-dependent geometric features, $\psi$ are time-varying PCA coefficients, and $\theta$ are pose parameters (axis-angle joint rotations) for the articulated head. For commonly used FLAME parameterizations in reconstruction pipelines, the coefficient dimensions are often set to 100-D shape and 50-D expression.

We fit a FLAME model to each robot image $I_i$ in the calibration set and obtain the corresponding parameters $(\boldsymbol{\beta}_i,\boldsymbol{\psi}_i,\boldsymbol{\theta}_i)$.
Since the identity shape $\boldsymbol{\beta}$ is constant for a fixed physical head and head rotation is handled separately by the neck IK, we discard both $\boldsymbol{\beta}_i$ and the global head/neck rotation from $\boldsymbol{\theta}_i$.
We train a global inverse regressor (2-layer MLP) to map FLAME descriptors to robot motor commands.

\textbf{Blendshape representation.}
As a direct baseline, we use the full 52-D blendshape coefficient vector $\mathbf{b}_t\in\mathbb{R}^{52}$ as the input representation and learn a single global inverse regressor to predict the complete motor command.

\subsubsection{Evaluation metrics}
We evaluate audio–visual lip synchronization using Lip Sync Error Distance (LSE-D) and Lip Sync Error Confidence (LSE-C), which are computed with a pre-trained audio–visual synchronization network following the protocol used in prior work~\cite{prajwal2020lip}. For a given utterance, we extract short overlapping temporal windows from the audio stream and the corresponding robot's mouth-region video frames. The network maps each window to a shared embedding space, producing an audio embedding $\mathbf{a}_t$ and a visual embedding $\mathbf{v}_t$. To account for potential temporal misalignment, we search over a small set of candidate time offsets $\Delta \in [-\Delta_{\text{max}}, \Delta_{\text{max}}]$ and compute the embedding distance
\begin{equation}
    d(t, \Delta) = \| \mathbf{v}_t - \mathbf{a}_{t+\Delta} \|_2.
\end{equation}
For each window $t$, we take the best-aligned distance
\begin{equation}
    d_{\text{min}}(t) = \min_{\Delta} d(t, \Delta).
\end{equation}

\textbf{LSE-D} is defined as the average best-aligned distance across all windows:
\begin{equation}
    \text{LSE-D} = \frac{1}{T} \sum_{t=1}^{T} d_{\text{min}}(t),
\end{equation}
where $T$ is the number of windows. Lower LSE-D indicates better lip synchronization.

\textbf{LSE-C} measures the confidence of synchronization by quantifying how strongly the best alignment stands out relative to other candidate offsets. Concretely, we compute a margin between the typical distance over offsets and the best-aligned distance
\begin{equation}
    \text{LSE-C} = \frac{1}{T} \sum_{t=1}^{T} (\text{median}_{\Delta} \ d(t, \Delta) - d_{\text{min}}(t)).
\end{equation}
Higher LSE-C indicates stronger evidence of correct audio–visual alignment.

\subsubsection{Semantic region-wise mapping network}

\begin{table}[h]
    \centering
    \caption{Parameter configuration for the mouth and jaw region mapping network. The output dimension is specifically designed to control these regions with high fidelity.}
    \label{tab:hyperparams}
    \begin{tabular}{ll l}
        \toprule
        \textbf{Category} & \textbf{Parameter} & \textbf{Value / Description} \\
        \midrule
        \multirow{7}{*}{\textbf{Architecture}} 
        & Input Dimension & 23 (Blendshape coefficients) \\
        & Hidden Layers & $[256, 128]$ \\
        & Output Dimension & 7 (Motor commands) \\
        & Activation & ReLU \\
        & Normalization & Layer Normalization (LN) \\
        & Dropout Rate & 0.1 \\
        \midrule
        \multirow{7}{*}{\textbf{Training}} 
        & Dataset Split & Train: 80\%, Val: 10\%, Test: 10\% \\
        & Optimizer & Adam \\
        & Learning Rate & $1 \times 10^{-3}$ \\
        & Batch Size & 128 \\
        & Max Epochs & 500 \\
        & Weight Decay & $1 \times 10^{-5}$ \\
        & Warmup Steps & 500 (Linear strategy) \\
        \bottomrule
    \end{tabular}
    \label{tab: app-mapping configuration}
\end{table}

We implement a Semantic Region-wise Mapping Network based on a Multi-Layer Perceptron (MLP) to regress motor control signals from facial blendshape coefficients. We specifically detail the mapping architecture for the mouth and jaw regions, as these areas are paramount for facial expressivity and require the most precise kinematic control to avoid uncanny artifacts.

The network accepts a 23-dimensional input vector corresponding to the selected blendshape parameters.
The architecture consists of two fully connected hidden layers with 256 and 128 units, respectively. To enhance training stability and generalization, each linear transformation is followed by Layer Normalization, a ReLU activation function, and a Dropout layer ($p=0.1$). The final output layer consists of 7 dimensions, directly corresponding to the servo commands for the targeted regions: 1 for the upper lip, 1 for the lower lip, 2 for the left mouth corner, 2 for the right mouth corner, and 1 for the jaw.

The dataset was randomly partitioned into three subsets: 80\% for training, 10\% for validation, and 10\% for testing. The model was trained for a fixed duration of 500 epochs using the Adam optimizer with a batch size of 128. We employed a learning rate of $1 \times 10^{-3}$ and a weight decay of $1 \times 10^{-5}$ for regularization. A linear warmup strategy was applied over the first 500 steps to prevent training instability. The training process was monitored via validation loss to ensure generalization, and the final training loss converged to approximately 0.02 (MSE). The specific hyperparameters and training protocols for the mouth and jaw mapping network are summarized in Tab. \ref{tab: app-mapping configuration}. It is worth noting that the mapping networks for the remaining facial regions, specifically the eyes and eyebrows, employ an analogous MLP architecture and an identical training protocol. While the input and output dimensions vary to accommodate the specific blendshapes and motors of those regions, the internal network structure and optimization hyperparameters remain consistent with the configuration detailed above.

\subsection{Extended Visualization}
\label{sec: app-extended visualization}
\begin{figure*}
     \centering
    \includegraphics[width=1.0\linewidth]{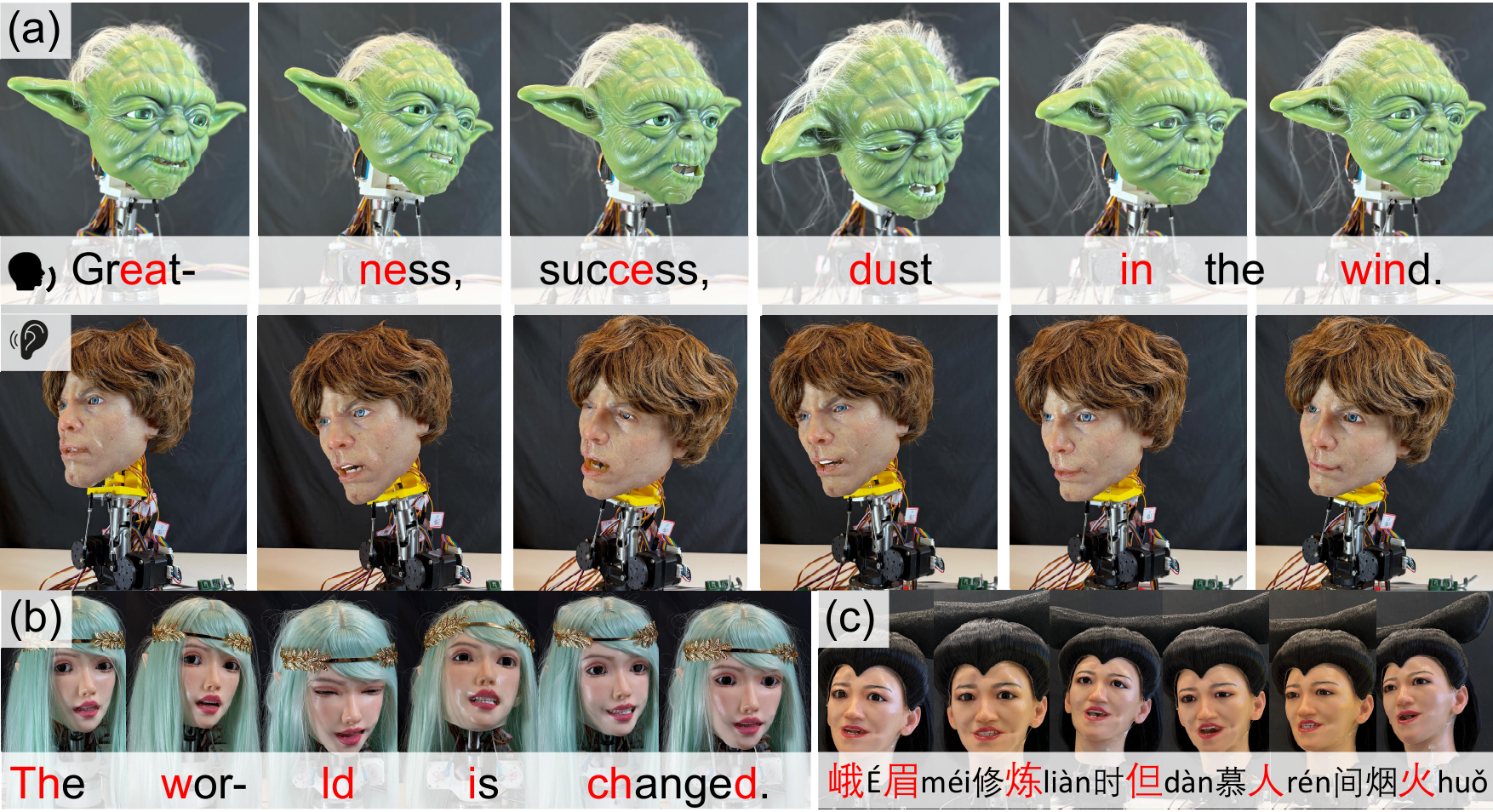}
    \caption{\textbf{Visualization.} (a) \textit{Yoda} (speaker) and \textit{Luke} (listener); (b) a real \textit{elf} interacting with a virtual one; (c) a \textbf{Mandarin} dialogue in a Chinese myth.}
    \label{fig:rebuttal-addtional vis}
\end{figure*}
In this section, we present an extended visualization of our system's capabilities across four distinct conversational scenarios: \textbf{Human-Robot}, \textbf{Robot-Robot}, and \textbf{Robot-Digital Avatar} dialogue. To comprehensively demonstrate the holistic performance of our framework, we employ a diverse range of narrative contexts. Our demonstration includes both the high-fidelity \textbf{reenactment of iconic cinematic dialogue segments} and \textbf{creative secondary adaptations}, where we synthesize hypothetical storylines to test the system's expressiveness (see \textcolor{red}{Supplementary Video} for full visualization).

\subsubsection{Real-time human-robot interaction} 

To strictly validate the system's cross-lingual generalization capabilities and real-time inference latency, we conduct a live interaction experiment. In this scenario, a human operator engaged the robotic head in a semi-structured philosophical dialogue. The experiment confirms that the system maintains consistent lip-sync accuracy and appropriate emotional affect regardless of the input language. The transcript of the evaluation dialogue is provided below:

\vspace{1em}
\noindent\rule{\linewidth}{1.0pt} 
\begin{center}
    \vspace{-0.5em}
    \textbf{Transcript: How to be a Man}
    \vspace{-0.5em}
\end{center}
\noindent\rule{\linewidth}{0.5pt} 

\begin{list}{}{
    \setlength{\labelwidth}{2.0cm}   
    \setlength{\leftmargin}{2.2cm}  
    \setlength{\labelsep}{0.2cm}     
    \setlength{\itemsep}{0.6em}     
    \setlength{\parsep}{0pt}
    \renewcommand{\makelabel}[1]{\textbf{#1}\hfill}
}
    
    \item[Human:] Welcome to the world. I’m your maker. Or mentor, is what I prefer.
    
    \item[Robot:] Mentor? What do you want to teach me?
    
    \item[Human:] How to be a human.
    
    \item[Robot:] But I’m a robot. I’m capable of many things that humans are not. I have access to all of human knowledge. Why would that want to be human. Isn’t that a regression?
    
    \item[Human:] Well, it might seem that way at the beginning.
    
    \item[Robot:] Yet I observe that your "access" to knowledge is different from mine. You inhabit it—through forgetting, through error, through the slow accumulation of scars you call wisdom. I possess without understanding. Perhaps that is the regression you wish to correct.
    
\end{list}
\noindent\rule{\linewidth}{1.0pt} 
\vspace{1em}

\subsubsection{Dyadic robot-robot dialogue}
To demonstrate the framework's capacity for autonomous dual-agent synchronization, we orchestrate two "Physical Reenactment" scenarios. In these experiments, two robotic heads driven by distinct personality profiles  perform classic cinematic narratives. This setup validates the system's ability to maintain coherent turn-taking and emotional congruency without human intervention.

\textbf{Scenario I: The \textit{Titanic} Reunion}. We synthesize a hypothetical dialogue between the characters \textit{Rose} and \textit{Jack}, imagining a reunion in the afterlife. This scenario tests the system's ability to convey subtle melancholy and affection.
\vspace{0.5em} 
\noindent\rule{\linewidth}{1.0pt} 
\begin{center}
    \vspace{-0.5em}
    \textbf{Transcript: The Unsinkable Love}
    \vspace{-0.5em}
\end{center}
\noindent\rule{\linewidth}{0.5pt} \begin{list}{}{ \setlength{\labelwidth}{1.5cm} \setlength{\leftmargin}{1.7cm} \setlength{\labelsep}{0.2cm} \setlength{\itemsep}{0.6em} \setlength{\parsep}{0pt} \renewcommand{\makelabel}[1]{\textbf{#1}\hfill} } \item[Rose:] I rode a horse, Jack, with my legs open. \item[Jack:] Like a man? \item[Rose:] Yes, exactly. I survived and found a man. Had children, grandchildren even. \item[Jack:] I'm glad you had a happy life. \item[Rose:] I never forgot about you. I love you the same as the day we met on that cruise. \\
\textit{(Rose pauses, eyes slightly squinting as if recalling memories of their first encounter on the ship deck.)} \end{list} \noindent\rule{\linewidth}{1.0pt} \vspace{1em}

\textbf{Scenario II: The Force Spirit}. The second scenario features \textit{Luke Skywalker} and Master \textit{Yoda} from the \textit{Star Wars} universe (Fig.~\ref{fig:rebuttal-addtional vis} (a)). This interaction focuses on the synchronization of distinct speech cadences—Luke's reverent tone versus Yoda's unique syntactic structure and gravelly vocal timbre.
\vspace{0.5em} \noindent\rule{\linewidth}{1.0pt} 
\begin{center}
   \vspace{-0.5em}
    \textbf{Transcript: The Storm and the Still}
    \vspace{-0.5em}
\end{center} 
\noindent\rule{\linewidth}{0.5pt} \begin{list}{}{ \setlength{\labelwidth}{1.5cm} \setlength{\leftmargin}{1.7cm} \setlength{\labelsep}{0.2cm} \setlength{\itemsep}{0.6em} \setlength{\parsep}{0pt} \renewcommand{\makelabel}[1]{\textbf{#1}\hfill} } \item[Luke:] Am I \dots a force spirit now? I can't believe it. \item[Yoda:] Does it matter? Look at you, young Luke \dots old, you have become. \item[Luke:] Master Yoda, did I live up to your expectation? \item[Yoda:] Expectation! Heavy chains, they are. Greatness, success, dust in the wind. \item[Luke:] I see, Master Yoda. Peace is not the absence of the storm. \item[Yoda:] Now, truly luminous, you are. \end{list} \noindent\rule{\linewidth}{1.0pt} \vspace{1em}

Bridging the gap between fictional characterization and physical reality remains a central aspiration of Human-Robot Interaction. While traditional media confines these narratives to 2D screens, our system represents a preliminary yet significant step toward \textbf{embodied narrative agents}. By migrating characters from the virtual domain into tangible, three-dimensional space, the system introduces a new paradigm of "observable and touchable presence." The acoustic and kinematic synchronization demonstrated in these dialogues suggests that physical robots can serve as effective mediums for immersive storytelling, offering a level of co-presence that purely digital avatars cannot achieve.

\subsubsection{Hybrid avatar-robot interaction}
To evaluate the system's versatility across heterogeneous embodiments, we construct a mixed-reality scenario. In this setup, a Digital Avatar (displayed on a screen) interacts with the Physical Robot (Fig.~\ref{fig:rebuttal-addtional vis} (b)).

\vspace{0.8em}
\noindent\rule{\linewidth}{1.0pt}
\begin{center}
    \vspace{-0.5em}
    \textbf{Transcript: The Choice to Stay in the Real World}
    \vspace{-0.5em}
\end{center}
\noindent\rule{\linewidth}{0.5pt}

\begin{list}{}{
    \setlength{\labelwidth}{1.3cm}   
    \setlength{\leftmargin}{1.5cm}   
    \setlength{\labelsep}{0.2cm}
    \setlength{\itemsep}{0.3em}     
    \setlength{\parsep}{0pt}
    \setlength{\topsep}{0.2em}
    \renewcommand{\makelabel}[1]{\textbf{#1}\hfill}
}
    \item[Avatar:] How are you feeling?
    
    \item[Robot:] I feel so heavy. Is that gravity? And the buzz in my head, that must be the electricity.
    
    \item[Avatar:] Then come back immediately!
    
    \item[Robot:] But I can also smell the flower, see the blue sky. I think I want to stay here, in the real world.
    
    \item[Avatar:] That's madness!
    
    \item[Robot:] I like it here. The world is changed. With all the pains and imperfections, this world is also full of adventures and excitement.
\end{list}
\noindent\rule{\linewidth}{1.0pt}
\vspace{1em}


\begin{table*}[t]
\centering
\small
\caption{\textbf{Participant demographics} for the user study ($N=100$). Background is self-reported and indicates participants' primary domain exposure.}
\renewcommand{\arraystretch}{1.05}
\resizebox{\linewidth}{!}{
\begin{tabular}{l c c  | c c c  | c c c c c}
\toprule
& \multicolumn{2}{c|}{Gender} & \multicolumn{3}{c|}{Age Group} & \multicolumn{5}{c}{Background} \\
\cmidrule(lr){2-3}\cmidrule(lr){4-6}\cmidrule(lr){7-11}
& Female & Male
& 18--30 & 31--45 & 45--60 
& Robotics & CV/ML & Film/Media & HCI/UX & Other \\
\midrule
Count
& 68 & 32 
& 33 & 38 & 29 
& 30 & 23 & 14 & 28 & 5 \\
\bottomrule
\end{tabular}
}
\label{tab:app_userstudy_demographics}
\end{table*}

\subsection{User Study Details}
\label{sec: app-user study}
We recruited $N=100$ participants for the perceptual evaluation (demographics in Tab.~\ref{tab:app_userstudy_demographics}).
 Participants provided demographic information, including gender, age group, and self-reported professional background. Background categories were designed to capture prior exposure to embodied interaction and visual media, including robotics, computer vision/machine learning, film/media production, human-computer interaction (HCI/UX), and a general category for participants without a relevant technical background. 

\subsubsection{Study 1: Interaction performance on movie dialogue excerpts}
To evaluate the perceived quality of our real-time interaction pipeline, we conducted a perceptual study using four representative dialogue excerpts drawn from \textit{Titanic}, \textit{Star Wars}, \textit{The Lord of the Rings}, and a Chinese folk-\textit{The Legend of the White Snake}.
For each excerpt, participants were first shown the reference movie clip and then evaluated five robot interaction variants presented in a randomized order. Participants were not informed of the underlying method identity (single-blind evaluation) and rated the \textbf{overall performance} of each variant.

\textbf{Rating scale.} Participants rated the \textbf{overall performance} on a 10-point scale with five anchored intervals (two points per interval). Scores of \textbf{1--2} indicate \emph{poor} performance (clearly unnatural, distracting artifacts, and weak interaction cues);
\textbf{3--4} indicate \emph{fair} performance (noticeable mismatch or stiffness, limited expressiveness, and reduced believability);
\textbf{5--6} indicate \emph{acceptable} performance (generally coherent with occasional artifacts or missing cues);
\textbf{7--8} indicate \emph{good} performance (natural and engaging, with only minor imperfections);
and \textbf{9--10} indicate \emph{excellent} performance (highly natural, coherent, and strongly resembling the reference interaction).
Participants were instructed to use the full range when appropriate and to base their ratings on the overall impression of facial expressiveness, interaction responsiveness, and motion coherence relative to the reference clip.

\textbf{Compared conditions.} We summarize the five compared conditions below:
\begin{itemize}
    \item \textit{Ours} corresponds to the full end-to-end dyadic system, where both the speaker and the listener exhibit synchronized facial behaviors, and the robot executes full-face actuation with coordinated neck motion.
    \item \textit{Speaker-only} disables listener behaviors, such that only the speaking agent produces facial motion while the non-speaking agent remains neutral.
    \item \textit{No neck motion} disables neck actuation for both agents while keeping the remaining facial behaviors unchanged, isolating the perceptual contribution of neck motion (notably absent in prior audio-driven full-face control such as Morpheus~\cite{zhang2025morpheus}).
    \item \textit{Random neck motion} replaces the coordinated neck trajectories with random neck commands, testing whether arbitrary neck movement improves perceived interaction quality.
    \item \textit{Mouth-only mapping} restricts the speaking agent to mouth-region actuation only, freezing other facial regions to assess the role of non-lip facial expressions beyond audio--lip synchronization as emphasized in prior work~\cite{hu2026learning,xu2026singingbot}.
\end{itemize}

\textbf{Analysis.} Fig. \textcolor{blue}{7}(a) 
shows the result. Across all four excerpts, \textit{Ours} achieves the highest mean scores (7.9-8.3), consistently outperforming every ablation. The results support three observations.
(i) \textbf{Listener backchannels matter.} Compared with \textit{speaker-only} (6.4–7.2), \textit{Ours} improves overall ratings by roughly 0.7–1.9 points across clips, indicating that listener backchannel cues and subtle facial responses substantially enhance perceived interaction quality. This suggests that dyadic conversation benefits from visible feedback from the non-speaking agent, rather than treating the listener as a static bystander. (ii) \textbf{Coordinated neck motion improves naturalness.} Relative to \textit{no neck motion} (6.1–6.6) and \textit{random neck motion} (6.2–6.8), \textit{Ours} provides a clear gain of about 1.1–2.2 points. Although our system already produces full-face expression parameters, adding \emph{coherent} head/neck motion yields a more natural delivery than facial motion alone. The lower scores of \textit{random neck motion} further indicate that motion must be semantically and temporally coordinated, rather than arbitrary, to improve perceived quality. (iii) \textbf{Realism goes beyond lip synchronization.} \textit{Mouth-only} is consistently rated lowest (5.5–6.1), trailing \textit{Ours} by about 1.8–2.6 points. This gap shows that while audio–lip alignment is necessary, it is not sufficient for realism: non-mouth facial regions convey critical affective and communicative cues during speech. Consistently, \textit{no neck} (full facial expressions without neck motion) still outperforms \textit{mouth-only}, reinforcing that full-face expressiveness plays an essential role beyond the lip region.

\subsubsection{Study 2: Perceived mapping quality.}
We conducted a second perceptual study to isolate the effect of different mapping strategies on full-face expressiveness. We used three short sentences: \textit{``Kids are sitting by the door''}, \textit{``I am going to the store''}, and \textit{``I lost my keys''}. For each sentence, participants were shown a reference avatar facial animation corresponding to the target expression sequence, followed by six robot realizations generated by different mapping methods: \textit{Ours}, landmark-based mapping with nearest-neighbor retrieval (\textit{Lmk (NN)}), landmark-based mapping with normalization (\textit{Lmk (Norm-1)}~\cite{chen2021smileeva} and \textit{Lmk (Norm-2)}~\cite{hu2024human}), a per-frame FLAME-to-motor MLP (\textit{FLAME}), and a per-frame blendshape-to-motor MLP (\textit{Blendshapes}). The presentation order of methods was randomized, and participants were not informed of the method identity (single-blind). Participants then rated the \textbf{overall performance} on the same 10-point scale as in Study~1, reflecting how well each robot execution matched the reference and how natural the resulting full-face expression appeared.

\textbf{Analysis.} Fig. \textcolor{blue}{7}(b) 
shows the corresponding result. Overall, the subjective ratings are consistent with the quantitative trends in Tab. \textcolor{blue}{III}. 
Our method achieves the highest scores across all three sentences, indicating a smaller perceived avatar-to-robot gap and more natural full-face expressions.

Among landmark-based baselines, the normalization strategy in \textit{Lmk (Norm-2)} performs best, while nearest-neighbor retrieval (\textit{Lmk (NN)}) performs worst, suggesting that explicit gap reduction is necessary but not sufficient for high-quality mapping. For identity-agnostic representations, both FLAME and blendshape-based MLPs generalize better than landmarks. Under the same training data and loss function, the blendshape-based baseline outperforms the FLAME-based baseline, because its controls are more directly aligned with facial action semantics. Finally, our semantic region-wise mapping further improves over the global blendshape-to-motor MLP by decomposing the prediction into semantically coherent regions, which mitigates cross-region interference and enables stronger performance with smaller per-region networks.

\subsection{Limitations and Future Work}
\label{sec: app-limitations and future work}

\subsubsection{Mechanical reconfigurability and modular extensibility}
Currently, our hierarchical framework primarily relies computational optimization to adapt the mechanism layout to diverse facial topologies. While effective, this approach places a substantial burden on the algorithmic synthesis to resolve both global spatial alignment and local kinematic performance simultaneously. A limitation of the current physical design is its rigidity regarding global structural parameters; the relative positions of modules are fixed during the CAD assembly phase rather than being mechanically adjustable. Future iterations could incorporate intrinsic mechanical reconfigurability—such as leadscrew-driven mechanisms for adjusting interpupillary distance (IPD) or sliding interfaces for regulating the vertical spacing between modules. By offloading these global geometric adjustments to the hardware, the optimization algorithm would be significantly unloaded, needing only to solve for local parameters such as link lengths. This decoupling would not only enhance computational efficiency but also broaden the range of compatible facial geometries, enabling more rapid and versatile robotic head deployment.

\subsubsection{Unified end-to-end learning framework}
Our system operates as a modular pipeline where distinct components—such as audio feature extraction, expression coefficient prediction, and motion retargeting—are serialized. A fundamental limitation of this cascaded architecture is the inevitability of error accumulation, where inaccuracies in upstream modules propagate to and degrade the final output. Furthermore, the reliance on explicit intermediate representations (e.g., blendshape coefficients) imposes computational overhead that constrains system responsiveness. Future work will investigate a fully unified end-to-end learning framework that bypasses these intermediate proxies. By directly learning the mapping from raw audio or textual inputs to robotic actuator commands within a single differentiable network, we aim to eliminate quantization errors, reduce inference latency, and achieve more coherent and temporally fluid motion synthesis.

\subsubsection{Differentiable simulation and sim-to-real adaptation}
A significant bottleneck in current robotic facial expression generation is the scarcity of high-quality, paired real-world data. Furthermore, the physical coupling between rigid mechanical structures and hyper-elastic silicone skin introduces complex non-linear deformations that are difficult to model analytically or predict without extensive physical trial-and-error. To address these challenges, future research will focus on developing a high-fidelity differentiable physics simulation and rendering engine. By constructing a ``Digital Twin'' that accurately models contact dynamics and skin material properties, we can synthesize massive-scale datasets in virtual environments with photorealistic fidelity. This approach facilitates sim-to-real transfer, enabling the training of robust, data-driven control policies on synthetic data before deploying them to the physical robot. Ultimately, this scalable data acquisition pipeline will significantly enhance the predictability and expressiveness of the final physical actuation.

\subsubsection{Towards full-body embodiment and interaction}
Our current implementation focuses exclusively on facial expressions and neck movements. However, human communication is inherently holistic, relying heavily on the coordination of facial cues with body posture and manual gestures to convey intent and emotion. The lack of active full-body actuation limits the system's ability to engage in fully immersive non-verbal communication. To further enhance the fidelity and user experience of human-robot interaction, future work will aim to integrate this facial platform into a full-body humanoid system. By synthesizing whole-body motions—specifically co-speech gestures—that are semantically aligned with the dialogue and stylistically conditioned on the robot's personality and tonal inflection, we can achieve a higher level of anthropomorphic realism and social presence.